\documentclass{article}

\usepackage{arxiv}

\usepackage[utf8]{inputenc} 
\usepackage[T1]{fontenc}    
\usepackage{hyperref}       
\usepackage{url}            
\usepackage{booktabs}       
\usepackage{amsfonts}       
\usepackage{nicefrac}       
\usepackage{microtype}      
\usepackage{lipsum}		
\usepackage{graphicx}
\usepackage{natbib}
\usepackage{doi}

\usepackage{amssymb}
\usepackage{color}
\usepackage{threeparttable}
\usepackage{setspace}
\usepackage{hhline}
\usepackage{bbding} 
\usepackage[table]{xcolor} 
\usepackage{subfigure}
\usepackage{graphicx}
\usepackage{booktabs}
\usepackage{amsmath}
\usepackage{multirow}
\usepackage{mathtools}
\usepackage{longtable}

\definecolor{C1}{HTML}{93BFCF}
\definecolor{C2}{HTML}{A0C3D2}
\definecolor{C3}{HTML}{BDCDD6}
\definecolor{C4}{HTML}{EEE9DA}
\definecolor{C5}{HTML}{FFF1DC}
\definecolor{C6}{HTML}{E8D5C4}
\definecolor{C7}{HTML}{EEEEEE}

\title{Multi-modal Multi-task Pre-training for Improved Point Cloud Understanding}

\author{{Liwen Liu$^1$, Weidong Yang$^1$, Lipeng Ma$^1$, Ben Fei$^2$}\\
	$^1$ Fudan University $^2$ The Chinese University of Hong Kong\\
	\texttt{liwenliu21@m.fudan.edu.cn, wdyang@fudan.edu.cn, lpma@m.fudan.edu.cn benfei@cuhk.edu.hk}}

\date{}

\renewcommand{\shorttitle}{\textit{arXiv} Template}

\hypersetup{
pdftitle={A template for the arxiv style},
pdfsubject={q-bio.NC, q-bio.QM},
pdfauthor={David S.~Hippocampus, Elias D.~Striatum},
pdfkeywords={First keyword, Second keyword, More},
}

\begin{document}
\maketitle

\begin{abstract}
Recent advances in multi-modal pre-training methods have shown promising effectiveness in learning 3D representations by aligning multi-modal features between 3D shapes and their corresponding 2D counterparts. However, existing multi-modal pre-training frameworks primarily rely on a single pre-training task to gather multi-modal data in 3D applications. This limitation prevents the models from obtaining the abundant information provided by other relevant tasks, which can hinder their performance in downstream tasks, particularly in complex and diverse domains. In order to tackle this issue, we propose MMPT, a Multi-modal Multi-task Pre-training framework designed to enhance point cloud understanding. Specifically, three pre-training tasks are devised: (i) Token-level reconstruction (TLR) aims to recover masked point tokens, endowing the model with representative learning abilities. (ii) Point-level reconstruction (PLR) is integrated to predict the masked point positions directly, and the reconstructed point cloud can be considered as a transformed point cloud used in the subsequent task. (iii) Multi-modal contrastive learning (MCL) combines feature correspondences within and across modalities, thus assembling a rich learning signal from both 3D point cloud and 2D image modalities in a self-supervised manner. Moreover, this framework operates without requiring any 3D annotations, making it scalable for use with large datasets. The trained encoder can be effectively transferred to various downstream tasks. To demonstrate its effectiveness, we evaluated its performance compared to state-of-the-art methods in various discriminant and generative applications under widely-used benchmarks.
\end{abstract}


\section{Introduction}
3D visual understanding has garnered significant attention in recent years owing to its increasing applications in augmented reality (AR), virtual reality (VR), autonomous driving, metaverse, and robotics~\citep{fei2022comprehensive, fei2023self,zhu2024advancements}.
The initial stage in point cloud understanding involves extracting discriminative geometric features, known as geometric representation learning (GRL). 
With adequate annotated data, GRL can be highly effective by integrating various neural networks, such as PointNet~\citep{qi2017pointnet}, PointNet++~\citep{qi2017pointnet++}, and DGCNN~\citep{wang2019dynamic}, to improve downstream tasks, such as classification and segmentation~\citep{zhang2024tcfap,xie20243d,xu2025visual}.
However, the process of collecting and annotating 3D data remains expensive and labor-intensive~\citep{yu2022point, huang2021spatio}.
While training on synthetic scans shows promise in alleviating the scarcity of labeled real-world data, GRL models trained in this manner are susceptible to domain shifts~\citep{he2022masked, zhang2022point}.

Self-supervised learning (SSL), as an unsupervised learning paradigm, provides a solution to the limitations of supervised models and has been successfully applied in 2D domains~\citep{chen2020simple,liu2025gs,liu2025gaussian2scene}. 
This has sparked a recent surge of interest in leveraging self-supervised learning to extract powerful features for 3D point clouds~\citep{fei2024parameter,fei2024curriculumformer,fei2024towards}. 
Most of the existing self-supervised learning methods adopt the encoder-decoder architecture, where the encoder's parameters are updated based on the decoder's reconstruction of point cloud objects~\citep{liu2022masked}.
However, these approaches face several challenges, including: 
i) Reconstructing 3D objects is not always feasible due to the discrete nature of point clouds.
ii) Unimodal losses, such as mean squared error and cross-entropy, are inadequate for capturing various geometric details in the original data.

To this end, researchers have explored other modalities that are more abundantly available, such as images, to provide additional supervisory signals for learning 3D representations~\citep{afham2022crosspoint,zhang2022pointclip}.
This approach has not only improved the ability to represent single-modal data, but has also facilitated the development of more comprehensive multi-modal representation capabilities~\citep{zhu2022pointclip}.
These efforts have shown promising outcomes and have partially alleviated the need for densely annotated single-modal data in the 3D domain.
However, these multi-modal pre-training methods still rely on a single pre-text task, which limits the acquisition of abundant information provided by other related pre-text tasks, ultimately hindering the performance of the pre-trained models for downstream tasks.

To tackle these challenges, we introduce a \textbf{M}ulti-modal \textbf{M}ulti-task \textbf{P}re-\textbf{T}raining framework, named \textbf{MMPT}, for self-supervised point cloud representation learning.
In detail, three pre-text tasks for pre-training are designed:
(i) The first pre-text task, \textbf{T}oken-\textbf{L}evel \textbf{R}econstruction (\textbf{TLR}), aims to recover masked tokens via cross-entropy, which is a commonly employed pre-training method for point cloud data.
As previously mentioned, while this pre-text task is effective for downstream tasks, reconstructing 3D objects can be challenging due to the discrete nature of point clouds, and cross-entropy loss is insufficient for learning detailed geometries.
To enhance the representative learning ability of the encoder, we combine the other two pre-text tasks.
(ii) \textbf{P}oint-\textbf{L}evel \textbf{R}econstruction (\textbf{PLR}) is designed to address the challenge of reconstructing point clouds due to their discrete nature.
Furthermore, the reconstructed point cloud from this pre-text task can be viewed as a transformed point cloud and can be utilized in the final task.
(iii) To improve the ability to capture detailed geometries besides cross-entropy loss, we introduce \textbf{M}ulti-modal \textbf{C}ontrastive \textbf{L}earning.
(\textbf{M}CL) consists of intra-modal learning and cross-modal learning.
After undergoing our multi-modal multi-task pre-training without manual annotation, we can transfer the trained encoder to various downstream tasks. We demonstrate our superior performance by comparing our method against widely used benchmarks.

The contributions of our MMPT can be summarized as follows:
\begin{itemize}
    \item We propose MMPT, a novel multi-modal and multi-task pre-training framework for improving point cloud understanding. This is the first time that multi-task pre-training has been integrated into 3D point cloud pre-training.
    \item Our MMPT framework comprises three pre-text tasks: token-level reconstruction, point-level reconstruction, and multi-modal contrastive learning. These tasks work in tandem to produce a powerful encoder that can be seamlessly transferred to downstream tasks with high effectiveness.
    \item We achieved comparable performance on five different downstream tasks, surpassing not only our competitors but also demonstrating improved generalization capability. Furthermore, we analyze the superiority of our approach by comparing it to existing self-supervised learning methods.
\end{itemize}

\section{Related Works}
\subsection{Self-supervised Learning on Point Clouds} 
Self-supervised learning (SSL) aims to extract robust and general features from unlabeled data, thereby mitigating the need for time-consuming data annotation and achieving superior performance in transfer learning tasks.

\textbf{Generative methods} learn features through self-reconstruction by encoding the point cloud into a feature or distribution and then decoding it back into the original point cloud~\citep{fei2024progressive,fei2025point}.
Recently, a wide variety of self-supervised methods based on Transformer architecture have been proposed.
For instance, Point-BERT~\citep{yu2022point} predicts discrete tokens, while Point-MAE~\citep{liu2022masked} randomly masks patches in the input point clouds and reconstructs the missing points.
An alternative to generative methods is to utilize generative adversarial networks for generative modeling.

\textbf{Discriminative methods} can learn point cloud representations by leveraging auxiliary handcrafted prediction tasks.
Jigsaw3D~\citep{sauder2019self} employs a 3D Jigsaw puzzle as a self-supervised learning task and utilizes contrastive techniques to train an encoder for downstream tasks. 
PointContrast~\citep{xie2020pointcontrast} introduces a pretext task that emphasizes maintaining consistent representations of a single point cloud from different viewpoints, with a focus on high-level scene understanding tasks. Based on this task, it investigates a unified comparative paradigm framework for 3D representation learning.
CrossPoint~\citep{afham2022crosspoint} combines information from both 3D and 2D modalities, emphasizing powerful shared features between them. Despite requiring challenging point cloud rendering outcomes, the approach is straightforward and efficient.
To facilitate contrastive learning tasks, Du et al.~\citep{du2021self} utilized self-similar point cloud patches from a single point cloud as positive or negative examples. In addition, they actively acquired hard negative examples in proximity to positive samples to enhance the discriminative feature learning process.
STRL~\citep{huang2021spatio}, which is an extension of BYOL to 3D point clouds, utilizes the interaction between online and target networks to learn representations.
By integrating the strengths of both generative and discriminative approaches, we propose a more comprehensive method for leveraging multi-task pre-training. This approach results in better representations, as it combines the benefits of both approaches.


\subsection{Multi-modal Representation Learning} 
This paper aims to leverage additional learning signals that are inherent to different modalities, such as 2D images, in addition to 3D point clouds.
These modalities contain rich contextual and textural information, as well as dense semantics. 
However, current methods in this field primarily focus on contrastive learning of global feature matching~\citep{afham2022crosspoint, fei2025multi,fei2022vq}.
To illustrate, a discriminative center loss is proposed by \citep{jing2021cross} to align features of point clouds, meshes, and images. 
Likewise, an intra- and inter-modal contrastive learning framework is presented by \citep{afham2022crosspoint} that operates on augmented point clouds and their corresponding 2D images. 
Another approach involves utilizing prior geometric information to establish dense associations and explore fine-grained local feature matching. For instance, Liu et al.~\citep{liu2021learning} proposed a contrastive knowledge distillation method to align fine-grained 2D and 3D features, while~\citep{li2022simipu} introduced a simple contrastive learning framework for inter- and intra-modal dense feature contrast, which employs the Hungarian algorithm to improve correspondence.
Recently, significant progress has been made by directly utilizing pre-trained 2D image encoders through supervised fine-tuning. For instance, Image2Point~\citep{xu2021image2point} suggests transferring pre-trained weights by inflating the convolutional layers. Meanwhile, P2P~\citep{wang2022p2p} proposes projecting 3D point clouds onto 2D images and feeding them into the image backbone via a learnable coloring module.

\subsection{Multi-task Pre-training}

Multi-task learning involves training models to predict multiple output domains from a single input.
A common technique in multi-task learning entails employing a solitary encoder to acquire a shared representation that subsequently passes through multiple task-specific decoders~\citep{ghiasi2021multi}.
In contrast to other approaches, our method incorporates multiple tasks in both the input and the output, accompanied by masking. 
Additionally, several studies have investigated the significance of task diversity in improving transfer performance~\citep{ghiasi2021multi}.
These studies suggest that learning solely from a single task is inadequate and that using a set of tasks can comprehensively encompass the wide range of potential downstream tasks in vision. 
Our MMPT leverages multiple tasks to acquire more generalized representations capable of addressing multiple downstream tasks.

\section{The Framework of MMPT}

\begin{figure*}[htb]
    \centering
    \includegraphics[width=1.0\textwidth,height=0.75\textwidth]{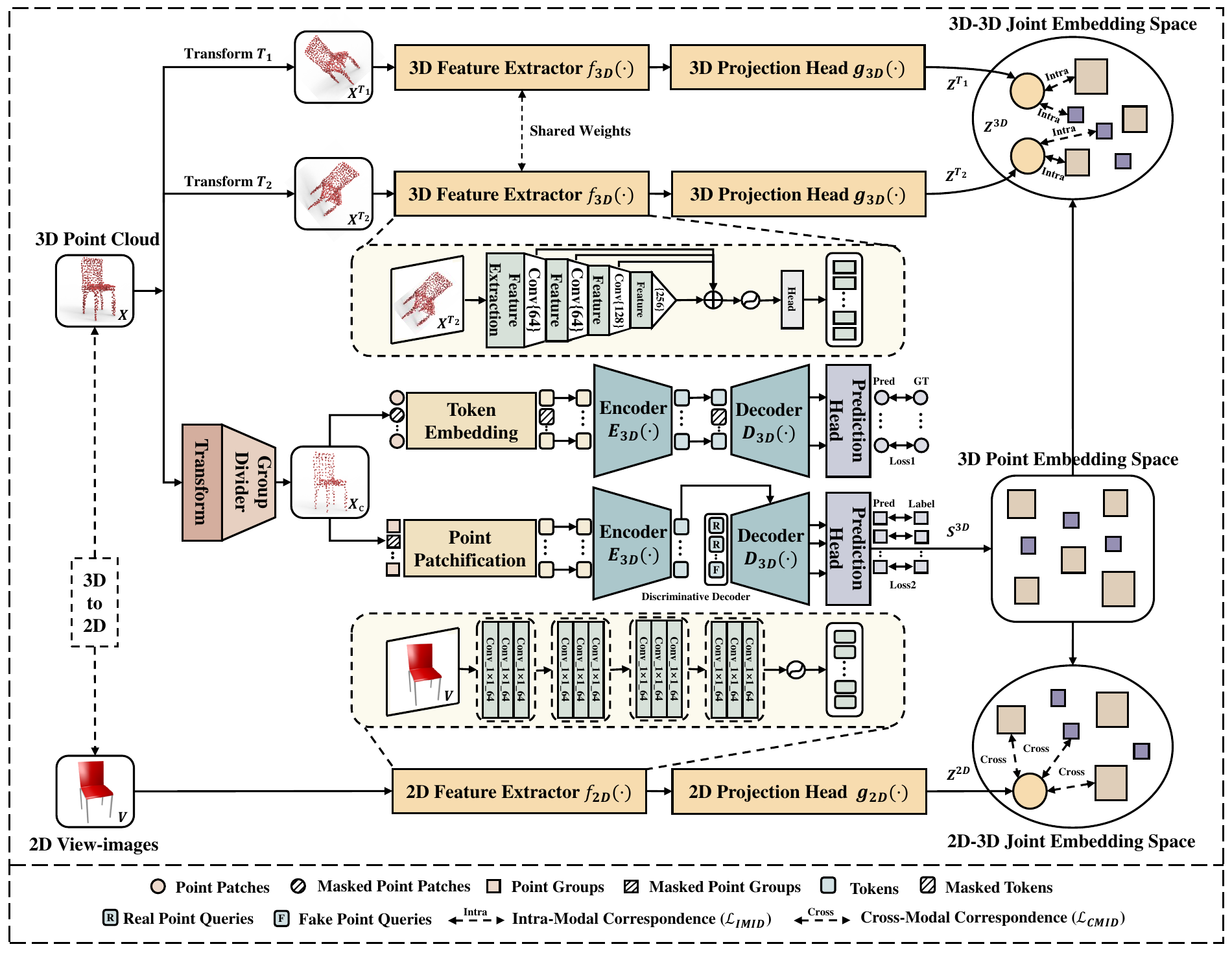}
    \caption{The overall framework of our MMPT. Our MMPT is a novel multi-task pre-training framework that consists of three different tasks: (i) Masked point tokens prediction task in TLR, which aims to recover masked tokens via cross-entropy. (ii) Masked point groups prediction task in PLR, which addresses the challenge of reconstructing point clouds due to their discrete nature. And (iii) 2D images-3D point clouds correspondence task in MCL. Upon completing our multi-modal multi-task pre-training without manual annotation, the trained encoder can be transferred to various downstream tasks.}
    \label{fig1:MMPT}
\end{figure*}

\subsection{Overview}
The overall framework of our MMPT is shown in Fig.~\ref{fig1:MMPT}. Our MMPT framework consists of three main pre-text tasks: \textbf{Masked Point Tokens Prediction Task} in TLR, \textbf{Masked Point Groups Prediction Task} in PLR, and \textbf{2D Images-3D Point Clouds Correspondence Task} in MCL. 
This section begins by introducing the masked point tokens prediction task, which enhances the transformer architecture's ability in classification, as explained in Section~\ref{sec:3.2}. We then introduce the masked point groups prediction task, which improves the backbone's capability for generating output, in Section~\ref{sec:3.3}. Finally, we provide details on the 2D images-3D point clouds correspondence network in Section~\ref{sec:3.4}.

\subsection{Masked Point Tokens Prediction Task in TLR}
\label{sec:3.2}

\begin{figure}[t]
    \centering    
    \includegraphics[width=0.5\linewidth]{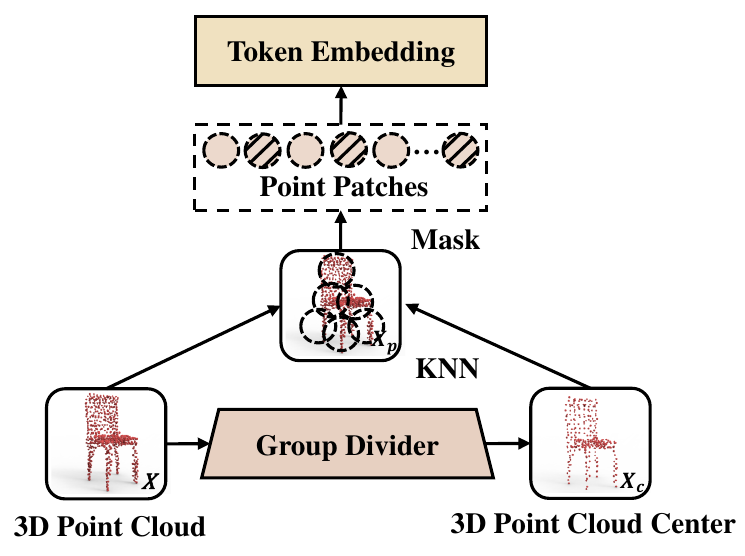}
    \caption{Illustrations of the masked point tokens prediction task, which enhances the classification capabilities of the transformer architecture.}
    \label{fig:2-1}
\end{figure}

\textbf{Masking and Embedding Stage.} 
Since a point cloud is a set of unordered points, grouping it into point patches has been shown to provide a better understanding and description of the local information of 3D shapes. 
As illustrated in Fig.~\ref{fig:2-1}, the masking and embedding stage aims to provide more accurate, detailed, and semantic point cloud data. 
In this stage, the input point cloud is divided into irregular point patches, after which these patches are randomly masked and embedded into tokens.

Specifically, suppose that a point cloud is $X\in R^{N\times3}$ with the size of $N\times3$ as input, we first adopt Furthest Point Sampling (FPS) to sample $M$ center points from the overall point cloud $X$ with a fixed sample ratio, which downsamples the point number from $N$ to $M$, denoted as $X_{c} = \left \{c_{i} \right \}_{i=1}^{M}\in R^{M\times3}$. Note that $X_{c} = \left \{c_{1}, c_{2},\ldots,c_{M}\right \} \in X$. To further form point patches by center points and their neighborhood points, the K-nearest neighborhood algorithm (KNN) is utilized to select a subset of K neighbors on each center point. By grouping M local patches, we generate point patches $X_{p} \in R^{M\times K\times3 }$ consisting of points location and surrounding geometric features. This process can be written as:
\begin{equation}
\begin{aligned}
&X_{c} = FPS\left (X\right)   \\
X_{p} &= KNN\left (X,X_{c}\right)  
\end{aligned}
\end{equation}

Then, given point patches and a masking ratio $\gamma \in \left ( 0,1 \right )$, we randomly mask a part of point patches $P_{mask}  \in R^{\gamma M\times K\times3 }$ and generate visible point patches $P_{vis}  \in R^{\left (1-\gamma \right )  M\times K\times3 }$. It is important to choose the masking strategy and the masking ratio due to their significant impact on the performance of mask transformers. For the masking strategy, we choose the random masking strategy, which can separately mask point patches as much as possible, keeping information complete by considering point patches overlap. For the masking ratio, we set the high masking ratio $\gamma = 0.8$ according to the experimental results, to better obtain latent representations from visible point patches $P_{vis}$.

After applying the random masking strategy to point patches, we adopt a mini-PointNet to implement instantiation of token embedding as the input to the encoder, which is composed of a multi-layer perceptron (MLP) and a max pooling layer. The initial tokens $T_{vis} \in R^{\left (1-\gamma \right )M\times D}$ is calculated as:
\begin{equation}
\begin{aligned}
T_{vis}= PointNet\left ( P_{vis}\right ) 
\end{aligned}
\end{equation}
In particular, the point cloud naturally has position information in 3D data. Since point patches are center normalized, appending the position embeddings of centers is essential. Following the prior study~\citep{yu2022point}, we use a small MLP network to learn position embeddings from center coordinates.

\textbf{Asymmetrical Autoencoder Stage.} In this stage, we adopt the strategy of shifting mask tokens to the decoder, which not only avoids giving out location information but also improves computation efficiency. Inspired by MAE~\citep{he2022masked}, we first adopt a standard transformer as the backbone network to construct the encoder with an asymmetric encoder-decoder design. During pre-training, the encoder takes the visible tokens $T_{vis}$ as input and adds positional embedding (PE) in each transformer block to provide details regarding the patch's location information. After $T_{vis}$ passes through the transformer backbone, we get the latent representation vector $T_{enc} \in R^{\left (1-\gamma \right )M\times D}$ (i.e. the global feature), where $D$ represents an embedded dimension. This process can be formulated as:
\begin{equation}
\begin{aligned}
T_{enc}= E_{3D}\left ( T_{vis},PE\right ) 
\end{aligned}
\end{equation}
Then the decoder is designed for outputting decoded mask tokens $H_{mask} \in R^{\gamma M\times D}$. Specifically, we also follow Point-MAE~\citep{pang2022masked} by utilizing a standard transformer with fewer blocks for construction. In the decoder, these encoded visible tokens $T_{enc}$, mask tokens $T_{mask}$, and their positional embeddings $PE$ are fed into the standard transformer. This process is defined as follows:
\begin{equation}
\begin{aligned}
H_{mask}=  D_{3D}\left (concat\left ( T_{enc},T_{mask}\right ),PE\right ) 
\end{aligned}
\end{equation}
Finally, the decoder output $H_{mask}$ is fed to a fully connected (FC) layer for reconstructing mask point patches $P_{pre} \in R^{\gamma M\times K\times3}$ as:
\begin{equation}
\begin{aligned}
P_{pre}= Reshape\left (FC\left ( H_{mask}\right )\right ) 
\end{aligned}
\end{equation}

\begin{figure}[t]
    \centering
    \includegraphics[width=0.5\linewidth]{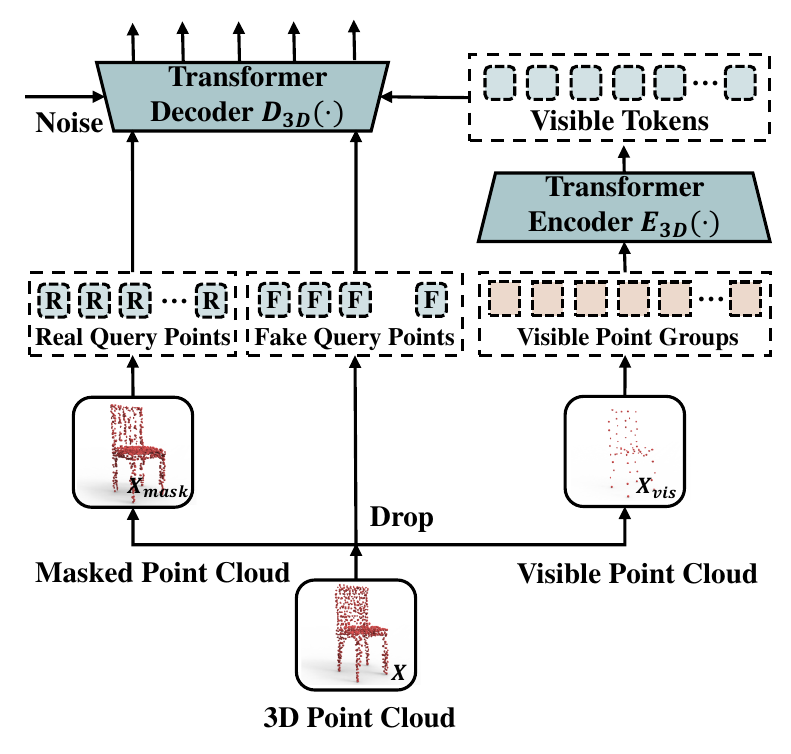}
    \caption{Illustrations of the masked point groups prediction task, which enhances the generation ability of backbone.}
    \label{fig:3}
\end{figure}

\subsection{Masked Point Groups Prediction Task in PLR}
\label{sec:3.3}

In the masked point groups prediction task, there are two main parts: the masked transformer and the discriminative decoder. The masked transformer is used to model the correlation between sparsely distributed unmasked groups, while the discriminative decoder assists the network in predicting the small number of visible point groups to determine the 3D shapes. 

\textbf{Grouping and Masking Stage.} As shown in Fig.~\ref{fig:3}, we first consider a 3D point cloud with $N$ points as the input $X\in R^{N\times3}$, which is downsampled by using FPS to produce patch centers. Then for each patch center, we find a subset of nearest neighbor points by applying the KNN, and form all these subsets as local groups $X_{g} \in R^{M\times K\times3}$. By randomly masking a proportion of them, we divide point groups into masked groups $X_{mask}$ and unmasked groups $X_{vis}$.

\textbf{Masked Transformer Stage.} During the masked transformer stage, the encoder takes visible local groups as input and outputs the global representations, which are composed of stacked multi-head self-attention layers (MSA) and a fully connected feed-forward network (FFN). Before being fed to the encoder, visible groups $X_{vis}$ are instantiated into group embeddings $T_{group}$ via a lightweight PointNet~\citep{qi2017pointnet} and converted into positional embeddings $T_{pos}$ via an MLP, respectively. 

Formally, we define the deep representations as the input embeddings $T_{input}$, which are the combination of these two embeddings $T_{group}$ and $T_{pos}$. Inspired by ViT, we also append a class token in front of the input sequences along the group dimension. Suppose the learnable class token is [CLS], which plays a crucial role in learning the overall structure of the point cloud and is applied to the downstream tasks. In this way, the overall input sequence of the Transformer can be expressed as $T_{input}=\{[CLS],t_1,t_2,\ldots,t_M\}\in R^{(M+1)\times D}$. After $T_{input}$ passes through $l$ layers of transformer blocks in the encoder network, we get the output of the last layer $T_{l}=\{[CLS]^{l},t_1^{l} ,t_2^{l},\ldots,t_M^{l}\}\in R^{(M+1)\times D}$, which indicates the encoded representations of the input groups with global receptive field.

\textbf{Discriminative Decoder Stage.} During the discriminative decoder stage, the decoder takes feature representations as input and outputs logits $S^{3D}$ and predicted queries $Q_{pre}$ through an MLP classification head. In particular, we denote a series of real queries $Q_{real}$ sampled from the masked groups and a series of fake queries $Q_{fake}$ sampled over the entire 3d space. Subsequently, the one-layer transformer decoder takes in them $\{Q_{real}+pos\}\cup\{Q_{fake}+pos\}$, and performs each query $q\in \{Q_{real},Q_{fake}\}$ through cross-attention $CA\left(q,t^{l}+pos\right)$ with the encoder outputs. Such a strategy that trains the decoder to distinguish between real and fake queries has two advantages: firstly, it helps the network to infer the 3D structure based on visible point groups in small amounts; secondly, it does not predict the coordinates of masked groups, thus preventing the leakage of position information.

\begin{figure}[t]
    \centering
    \includegraphics[width=0.5\linewidth]{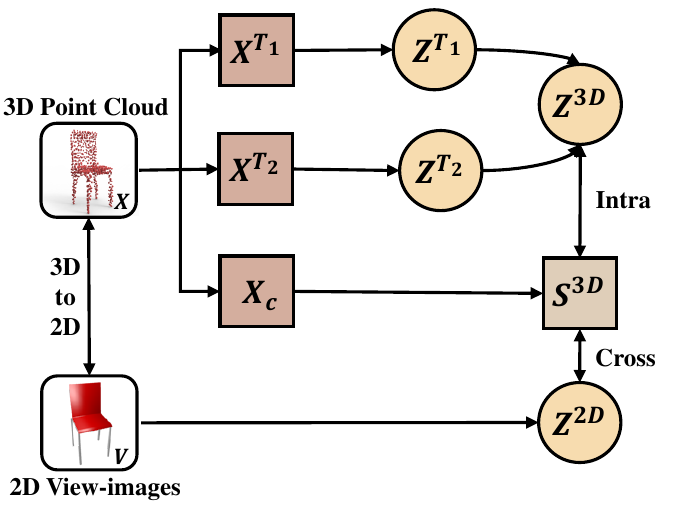}
    \caption{The pipeline of MCL, which can improve the classification ability of the network.}
    \label{fig:4}
\end{figure}

\subsection{2D Images-3D Point Clouds Correspondence Task in MCL}
\label{sec:3.4}

To enhance our understanding of 3D point clouds, we learn transferable representations in a self-supervised manner from 3D point clouds and 2D images, upon recent advances in intra-modal learning~\citep{chen2020simple} and cross-modal learning~\citep{he2022masked,pang2022masked,zhang2022point}.

\textbf{Intra-Modal Learning.} 
The goal of intra-modal learning is to encourage different projected vectors of the same point cloud to be similar while being dissimilar to projected vectors of other point clouds. We apply common 3D transformations during IMID, including scaling, rotation, normalization, elastic distortion, translation, and point dropout.

As illustrated in Fig~\ref{fig:3}, it takes as input two transformed versions of $X^{T_{1}}=\left \{ x_{i}^{t_{1}}   \right \}_{i=1}^{N} $ and $X^{T_{2}}=\left \{ x_{i}^{t_{2}}   \right \}_{i=1}^{N} $, which are randomly obtained by applying sequential combinations of transformations $T$ to the input point cloud $X$. To produce feature embeddings of transformed versions and construct feature vectors $Z^{T}$ in the invariant space, we employ the feature extractor $f_{3D}\left(\cdot  \right)$ and the projection head $g_{3D}\left(\cdot\right)$ in a successive manner. Since both transformed versions are utilized in point cloud projection, they share the weights of the 3D feature extractor. As our learning objective, we minimize the relative distance between the mean transformed version and the 3D logits ${S}^{3D}$, adjusting the dynamic range using a hyper-parameter $\tau$ through NT-Xent loss~\citep{chen2020simple}, which is defined as:
\begin{equation}\small
\label{eq:imid}
\begin{aligned}
& \ell_{i,{z}^{3D},{s}^{3D}}= \\
& -\log\frac{\exp(sim(\boldsymbol{z}_i^{3D},\boldsymbol{s}_i^{3D})/\tau)}{\sum\limits_{k=1\atop k\neq i}^N\exp(sim(\boldsymbol{z}_i^{3D},\boldsymbol{z}_k^{3D})/\tau)+\sum\limits_{k=1}^N\exp(sim(\boldsymbol{z}_i^{3D},\boldsymbol{s}_k^{3D})/\tau)} \\
& {z}_i^{3D}=\dfrac{1}{2}\left({z}_i^{t_1}+{z}_i^{t_2}\right),Z^{T} = g_{3D}\left(f_{3D}\left (  X^{T}\right )\right ) 
\end{aligned}
\end{equation}
where $\tau$ is the temperature hyper-parameter that controls the smoothness of the output distribution, ${z}_i^{3D}$ denotes the mean projected vector of the point cloud $X$.

\textbf{Cross-Modal Learning.} 
The goal of cross-modal learning is to leverage the implicit geometric and semantic correlation between 2D images and 3D point clouds, thus assisting 3D representation learning. In contrast to the sparse and irregular point clouds, 2D images can provide both fine-grained geometries and high-level semantics.

Concretely, for each rendered 2D image $V=\left \{ v_{i}\right \}_{i=1}^{N} $ of the point cloud $X$, we denote the visual backbone that maps the input to the feature space as $f_{2D}\left(\cdot\right)$. On top of 2D feature vectors, an image projection head $g_{2D}\left(\cdot\right)$ is utilized to project them into invariant space as vectors $Z^{2D}$. To do so, we employ contrastive learning to ensure that the similarity of the projected vector ${S}^{3D}$ and $Z^{2D}$ maps to nearby points while all the other projected vectors map to distant points. The contrastive aligned objective $L_{3D,2D}$ across point clouds and images is calculated with ${S}^{3D}$ and $Z^{2D}$ as:
\begin{equation}\small
\label{eq:cmid}
\begin{aligned}
& \ell_{i,{s}^{3D},{z}^{2D}}= \\
&-\log\frac{\exp(sim(\boldsymbol{s}_i^{3D},\boldsymbol{z}_i^{2D})/\tau)}{\sum\limits_{k=1\atop k\neq i}^N\exp(sim(\boldsymbol{s}_i^{3D},\boldsymbol{s}_k^{3D})/\tau)+\sum\limits_{k=1}^N\exp(sim(\boldsymbol{s}_i^{3D},\boldsymbol{z}_k^{2D})/\tau)} \\
&sim({s}_i^{3D},{z}_i^{2D})={s}_i^{3D\top} {z}_i^{2D}/\|{s}_i^{3D}\|\|{z}_i^{2D}\|,Z^{2D} = g_{2D}\left(f_{2D}\left (V\right )\right )
\end{aligned}
\end{equation}
where $\tau$ is the temperature coefficient, $sim\left(\cdot  \right)$ is a function that represents the dot product between the $L_2$-normalized vectors ${s}_i^{3D}$ and ${z}_i^{2D}$. 

\subsection{Loss Function}

\begin{table}[t]\footnotesize
\centering
\tabcolsep=0.31cm
\caption{\textbf{Shape classification on ModelNet40.} [ST] and [T] are denoted as the standard Transformers and Transformer-based methods, respectively. It is worth noting that the abbreviation 'Rep.' in the table indicates that we reproduced the results using the official codes.}
{%
\begin{tabular}{l|l|c}
\hline \toprule[1.5pt] \rowcolor{C7!70} 
                                 & Methods                  & Accuracy \\  \midrule[1pt]
\multirow{10}{*}{Supervised}     & PointNet~\citep{qi2017pointnet}                 & 89.2     \\ \hhline{~|-|-}  
                                 & \cellcolor{C7!70}PointNet++~\citep{qi2017pointnet++}             &\cellcolor{C7!70} 90.7     \\ \hhline{~|-|-} 
                                 & PointWeb~\citep{zhao2019pointweb}                 & 92.3     \\ \hhline{~|-|-} 
                                 & \cellcolor{C7!70}SpiderCNN~\citep{xu2018spidercnn}                  & \cellcolor{C7!70}92.4     \\ \hhline{~|-|-}  
                                 & PointCNN~\citep{li2018pointcnn}                 & 92.5     \\ \hhline{~|-|-} 
                                 & \cellcolor{C7!70}KPConv~\citep{thomas2019kpconv}                   & \cellcolor{C7!70}92.9     \\ \hhline{~|-|-}  
                                 & DGCNN~\citep{wang2019dynamic}                  & 92.9     \\ \hhline{~|-|-}  
                                 & \cellcolor{C7!70}RS-CNN~\citep{rao2020global}                  &\cellcolor{C7!70} 92.9     \\ \hhline{~|-|-} 
                                 & DensePoint~\citep{liu2019densepoint}                  & 93.2     \\ \hhline{~|-|-} 
                                 & \cellcolor{C7!70}PCT~\citep{guo2021pct}               & \cellcolor{C7!70}93.2     \\ \hhline{~|-|-}  
                                 & PVT~\citep{zhang2108pvt}             & 93.6     \\ \hhline{~|-|-}  
                                 & \cellcolor{C7!70}PointTransformer~\citep{zhao2021point}  & \cellcolor{C7!70}93.7     \\ \hhline{~|-|-}  
                                 & Transformer~\citep{yu2022point}     & 91.4     \\  \midrule[1pt]
\multirow{8}{*}{Self-supervised} & \cellcolor{C7!70}OcCo~\citep{wang2021unsupervised}                     & \cellcolor{C7!70}93.0     \\ \hhline{~|-|-} 
                                 & STRL~\citep{huang2021spatio}                      & 93.1     \\ \hhline{~|-|-}  
                                 & \cellcolor{C7!70} \begin{tabular}[c]{@{}c@{}}  Transformer\\ +OcCo~\citep{wang2021unsupervised} \end{tabular} & \cellcolor{C7!70}92.1     \\ \hhline{~|-|-}  
                                 & Point-BERT~\citep{yu2022point}        & 93.2    \\ 
                                 \hhline{~|-|-} 
                                 & \cellcolor{C7!70}Point-MAE~\citep{pang2022masked}      & \cellcolor{C7!70}93.8     \\
                                 \hhline{~|-|-}  
                                 & Point-MAE (Rep.)        & 93.1 \\ 
                                 \hhline{~|-|-} 
                                 & \cellcolor{C7!70}\textbf{MMPT}      & \cellcolor{C7!70}\textbf{93.9}     \\  \bottomrule[1.5pt]
\end{tabular}%
}
\label{table:modelnet40}
\end{table}

\begin{table}[t]\footnotesize
\centering
\caption{\textbf{The comparison of shape classification performance on ScanObjectNN.} The accuracy ($\%$) on three splits settings of ScanObjectNN are listed, where [S] stands for the fine-tuning model after self-supervised learning.}
{%
\begin{tabular}{l|ccc}
\toprule[1.5pt]
\rowcolor[HTML]{EFEFEF} Methods          & OBJ-BG & OBJ-ONLY & PB-T50-RS \\ \midrule[1pt]
PointNet~\citep{qi2017pointnet}           & 73.3   & 79.2     & 68.0      \\
\rowcolor[HTML]{EFEFEF}  PointNet++~\citep{qi2017pointnet++}      & 82.3   & 84.3     & 77.9      \\
DGCNN~\citep{wang2019dynamic}             & 82.8   & 86.2     & 78.1      \\
\rowcolor[HTML]{EFEFEF}  PointCNN~\citep{li2018pointcnn}          & 86.1   & 85.5     & 78.5      \\
SpiderCNN~\citep{xu2018spidercnn}            & 77.1   & 79.5     & 73.7      \\
\rowcolor[HTML]{EFEFEF}  BGA-DGCNN~\citep{uy2019revisiting}          & -   & -     & 79.7      \\
BGA-PN++~\citep{uy2019revisiting}             & -  & -   & 80.2      \\
\midrule[1pt]
Transformer~\citep{yu2022point}     & 79.9  & 80.6    & 77.2     \\
\rowcolor[HTML]{EFEFEF}  Transformer+OcCo~\citep{wang2021unsupervised}  & 84.9  & 85.5    & 78.8     \\
Point-BERT~\citep{yu2022point}      & 87.4  & 88.1    & 83.0     \\ \midrule[1pt]
\rowcolor[HTML]{EFEFEF}  MMPT      & \textbf{90.5}  & \textbf{91.0}    & \textbf{86.4}     \\ \bottomrule[1.5pt]
\end{tabular}%
}
\vspace{-0.4cm}
\label{table:scanobjectnn}
\end{table}

\begin{table}[htbp]\footnotesize
\tabcolsep=0.1cm
\centering
\caption{\textbf{The comparison of few-shot classification performance on ModelNet40.} For a fair comparison, the average accuracy ($\%$) and standard deviation ($\%$) of 10 experiments are reported.}
{%
\begin{tabular}{l|cc|cc}
\hline \toprule[1.5pt]
\multirow{2}{*}{Methods}                                           & \multicolumn{2}{c|}{5-way}                   & \multicolumn{2}{c}{10-way}                   \\ \cline{2-5}
                                                                   & \multicolumn{1}{c|}{10-shot}    & 20-shot     & \multicolumn{1}{c|}{10-shot}    & 20-shot    \\ \midrule[1pt]
DGCNN~\citep{wang2019dynamic}                                                             & \multicolumn{1}{c|}{91.8 $\pm$ 3.7} & 93.4 $\pm$ 3.2 & \multicolumn{1}{c|}{86.3 $\pm$ 6.2} & 90.9 $\pm$ 5.1 \\
\rowcolor[HTML]{EFEFEF} \begin{tabular}[c]{@{}c@{}}{[}S{]} DGCNN\\ +OcCo~\citep{wang2021unsupervised} \end{tabular}       & \multicolumn{1}{c|}{91.9 $\pm$ 3.3} & 93.9 $\pm$ 3.1 & \multicolumn{1}{c|}{86.4 $\pm$ 5.4} & 91.3 $\pm$ 4.6 \\ \midrule[1pt]
Transformer~\citep{yu2022point}                                                     & \multicolumn{1}{c|}{87.8 $\pm$ 5.2} & 93.3 $\pm$ 4.3 & \multicolumn{1}{c|}{84.6 $\pm$ 5.5} & 89.4 $\pm$ 6.3 \\
\rowcolor[HTML]{EFEFEF} \begin{tabular}[c]{@{}c@{}}{[}S{]}  Transformer\\ +OcCo~\citep{wang2021unsupervised} \end{tabular} & \multicolumn{1}{c|}{94.0 $\pm$ 3.6} & 95.9 $\pm$ 2.3 & \multicolumn{1}{c|}{89.4 $\pm$ 5.1} & 92.4 $\pm$ 4.6 \\
{[}S{]}Point-BERT~\citep{yu2022point}                                                  & \multicolumn{1}{c|}{94.6 $\pm$ 3.1} & 96.3 $\pm$ 2.7 & \multicolumn{1}{c|}{92.3 $\pm$ 4.5} & 92.7 $\pm$ 5.1 \\
\rowcolor[HTML]{EFEFEF} {[}S{]}MaskPoint~\citep{liu2022masked}                                                  & \multicolumn{1}{c|}{95.0 $\pm$ 3.7} & 97.2 $\pm$ 1.7 & \multicolumn{1}{c|}{91.4 $\pm$ 4.0} & 93.4 $\pm$ 3.5
\\
{[}S{]}Point-MAE~\citep{pang2022masked}                                                  & \multicolumn{1}{c|}{96.3 $\pm$ 2.5} & 97.8 $\pm$ 1.8 & \multicolumn{1}{c|}{92.6 $\pm$ 4.1} & 95.0 $\pm$ 3.0 \\\midrule[1pt]
\rowcolor[HTML]{EFEFEF} {[}S{]}MMPT                                                & \multicolumn{1}{c|}{\textbf{96.7 $\pm$ 2.7 }} & \textbf{97.9 $\pm$ 2.1} & \textbf{92.7 $\pm$ 4.3} & \textbf{95.7 $\pm$ 2.9} \\  \bottomrule[1.5pt]
\end{tabular}%
}
\label{table:fewshot}
\end{table}

\begin{table*}[t]\footnotesize
\centering
\tabcolsep=0.09cm
\caption{\textbf{The comparison of part segmentation performance on the ShapeNetPart.} The
mean IoU across all instance mIoU ($\%$) and the IoU ($\%$) for each categories are compared.}
\resizebox{\textwidth}{!}
{%
\begin{tabular}{l|c|cccccccccccccccc}
\hline \toprule[1.5pt]
Methods          & mIoU$_I$ & Aero & Bag  & Cap  & Car  & Chair & Ear  & Guitar & Knife & Lamp & Lap  & Motor & Mug  & Pistol & Rock & Skate & table \\ \midrule[1pt]
\rowcolor[HTML]{EFEFEF} PointNet~\citep{qi2017pointnet}          & 83.7  & 83.4 & 78.7 & 82.5 & 74.9 & 89.6  & 73.0 & 91.5   & 85.9  & 80.8 & 95.3 & 65.2  & 93.0   & 81.2   & 57.9 & 72.8  & 80.6  \\
\rowcolor[HTML]{FFFFFF} PointNet++~\citep{qi2017pointnet++}       & 85.1  & 82.4 & 79.0   & 87.7 & 77.3 & 90.8  & 71.8 & 91.0     & 85.9  & 83.7 & 95.3 & 71.6  & 94.1 & 81.3   & 58.7 & 76.4  & \textbf{82.6}  \\
\rowcolor[HTML]{EFEFEF} DGCNN~\citep{wang2019dynamic}            & 85.2  & 84.0   & 83.4 & 86.7 & 77.8 & 90.6  & 74.7 & 91.2   & 87.5  & 82.8 & 95.7 & 66.3  & 94.9 & 81.1   & \textbf{63.5} & 74.5  & \textbf{82.6}  \\ \midrule[1pt]
\rowcolor[HTML]{FFFFFF} Transformer~\citep{yu2022point}       & 85.1  & 82.9 & 85.4 & 87.7 & 78.8 & 90.5  & \textbf{90.8} & 91.1   & 87.7  & 85.3 & \textbf{95.6} & 73.9  & 94.9 & 83.5   & 61.2 & 74.9  & 80.6  \\
\rowcolor[HTML]{EFEFEF} Transformer+OcCo~\citep{wang2021unsupervised}  & 85.1  & 83.3 & 85.2 & 88.3 & \textbf{79.9} & 90.7  & 74.1 & 91.9   & 87.6  & 84.7 & 95.4 & 75.5  & 94.4 & 84.1   & 63.1 & 75.7  & 80.8  \\
\rowcolor[HTML]{FFFFFF} Point-BERT~\citep{yu2022point}       & 85.6  & 84.3 & 84.8 & \textbf{88.0} & 79.8 & 91.0  & 81.7 & 91.6   & 87.9  & 85.2 & \textbf{95.6} & 75.6  & 94.7 & 84.3   & 63.4 & 76.3  & 81.5  \\ \midrule[1pt]
\rowcolor[HTML]{EFEFEF} MMPT      & \textbf{86.5}  & \textbf{84.8} & \textbf{85.8} & 83.9 & 78.0 & \textbf{92.3}  & 77.1 &  \textbf{92.4}  & \textbf{88.9} & \textbf{85.9} & 95.5 & \textbf{75.4}  & \textbf{95.4}  & \textbf{84.6}  & \textbf{63.6} & \textbf{77.5}  & \textbf{82.6}  \\ \bottomrule[1.5pt]
\end{tabular}%
}
\label{table:partseg}
\end{table*}

\begin{table}[t]
\tabcolsep=0.35cm
\centering
\caption{
Semantic segmentation performance for Area 5 of the S3DIS dataset is compared. The evaluation metrics consist of mean accuracy (mAcc) and mean Intersection over Union (mIoU) calculated across all categories. Two different types of input features are employed: xyz, representing point cloud coordinates, and xyz+rgb, which combines coordinates with RGB information.
}
\begin{tabular}{l|ccc}
\toprule[1.5pt]
Methods                       & Input     & mAcc ($\%$) & mIoU ($\%$) \\ \midrule[1pt]
PointNet~\citep{qi2017pointnet}                     & xyz + rgb & 49.0      & 41.1      \\
\rowcolor[HTML]{EFEFEF}
PointNet++~\citep{qi2017pointnet++}                    & xyz + rgb & 67.1      & 53.5      \\
PointCNN~\citep{li2018pointcnn}                      & xyz + rgb & 63.9      & 57.3      \\
\rowcolor[HTML]{EFEFEF}
PCT~\citep{guo2021pct}                           & xyz + rgb & 67.7      & 61.3      \\ 
Transformer~\citep{yu2022point}                   & xyz       & 68.6      & 60.0      \\
\rowcolor[HTML]{EFEFEF}
Point-BERT~\citep{yu2022point}                     & xyz       & 69.7      & 60.5      \\ 
Point-MAE~\citep{pang2022masked}                    & xyz       & 69.9      & 60.8      \\ \midrule[1pt]
\rowcolor[HTML]{EFEFEF}
\textbf{MMPT}              & xyz       & \textbf{70.8}         & \textbf{62.5}        \\ \bottomrule[1.5pt]
\end{tabular}
\label{tab:indoorseg}
\end{table}

\begin{table}[ht]
\caption{
The comparisons of 3D object detection are presented based on the validation set of ScanNet V2. Our MMPT and Point-BERT use 3DETR as the backbone, while other approaches employ VoteNet for fine-tuning. Only geometric information serves as input for the subsequent task. The ``Input'' column specifies the input type used in the pre-training phase, with ``xyz'' denoting geometric information.
}
\centering
{
\begin{tabular}{l|cccc}
\toprule[1.5pt]
Methods                      & SSL & Pre-trained Input & $\textit{AP}_{25}$ & $\textit{AP}_{50}$ \\ \midrule[1pt]
\rowcolor[HTML]{EFEFEF}
VoteNet~\citep{qi2019deep}                      &     & -                 & 58.6 & 33.5 \\
STRL~\citep{huang2021spatio}                         & \CheckmarkBold   & xyz               & 59.5 & 38.4 \\
\rowcolor[HTML]{EFEFEF}
Implicit Autoencoder~\citep{yan2023implicit}         & \CheckmarkBold   & xyz               & 61.5 & 39.8 \\
RandomRooms~\citep{rao2021randomrooms}                  & \CheckmarkBold   & xyz               & 61.3 & 36.2 \\
\rowcolor[HTML]{EFEFEF}
PointContrast~\citep{xie2020pointcontrast}                & \CheckmarkBold   & xyz               & 59.2 & 38.0 \\
DepthContrast~\citep{wang2021unsupervised}                & \CheckmarkBold   & xyz               & 61.3 & -    \\
\rowcolor[HTML]{EFEFEF}
3DETR~\citep{misra2021end}                        &     & -                 & 62.1 & 37.9 \\
Point-BERT~\citep{yu2022point}                   & \CheckmarkBold   & xyz               & 61.0 & 38.3 \\
\rowcolor[HTML]{EFEFEF}
MaskPoint~\citep{liu2022masked}                    & \CheckmarkBold   & xyz               & 63.4 & 40.6 \\
Point-MAE~\citep{pang2022masked}                    & \CheckmarkBold   & xyz               & 63.0 & 42.4 \\
\rowcolor[HTML]{EFEFEF}
\textbf{MMPT} & \CheckmarkBold   & xyz               & \textbf{63.7}   & \textbf{42.8}    \\
\bottomrule[1.5pt]
\end{tabular}
}
\label{tab:indoordet}
\end{table}

\begin{table*}[t]\footnotesize
\centering
\tabcolsep=0.3cm
\caption{Quantitative comparisons on the PCN dataset for
shape completion (16,384 points) using $\ell1$ chamfer distance $\times 10^{3}$. Lower is better.}
\label{tab:PCN}
\resizebox{\textwidth}{!}
{%
\begin{tabular}{c|ccccccccc}
\toprule[1.5pt]
CD-$\ell1$($\times 10^{3}$) & Airplane & Cabinet & Car & Chair & Lamp & Sofa & Table & watercraft & Avg. \\ \midrule[1pt]
\rowcolor[HTML]{EFEFEF} 
ASFM~\citep{xia2021asfm} & 8.792 & 17.801 & 13.734 & 16.304 & 16.028 & 17.615 & 16.544 & 12.456 & 14.909 \\
CRN~\citep{wang2021cascaded} & 7.892 & 14.228 & 12.750 & 13.345 & 13.678 & 15.288 & 11.520 & 11.088 & 12.474 \\
\rowcolor[HTML]{EFEFEF} 
ECG~\citep{pan2020ecg} & 6.069 & 11.151 & 8.986 & 10.697 & 10.076 & 12.780 & 9.114 & 8.171 & 9.631 \\
FoldingNet~\citep{yang2018foldingnet} & 8.283 & 14.239 & 11.099 & 14.592 & 13.985 & 14.634 & 12.800 & 12.302 & 12.742 \\
\rowcolor[HTML]{EFEFEF} 
GRNet~\citep{xie2020grnet} & 9.433 & 14.928 & 12.019 & 14.523 & 12.166 & 15.424 & 12.782 & 11.017 & 12.786 \\
PCN~\citep{yuan2018pcn} & 6.825 & 12.948 & 9.953 & 13.174 & 13.366 & 14.434 & 11.452 & 10.488 & 11.580 \\
\rowcolor[HTML]{EFEFEF} 
TopNet~\citep{tchapmi2019topnet} & 7.785 & 14.939 & 11.643 & 15.449 & 14.037 & 15.264 & 12.297 & 12.315 & 12.966 \\
PoinTr~\citep{yu2021pointr} & 8.974 & 13.920 & 11.282 & 13.886 & 12.500 & 15.075 & 11.085 & 10.963 & 12.211 \\
\rowcolor[HTML]{EFEFEF} 
SnowflakeNet~\citep{xiang2021snowflakenet} & 5.262 & 10.372 & 8.847 & 9.103 & 7.717 & 10.714 & 7.663 & 7.221 & 8.362 \\
\midrule[1pt]
MMPT & \textbf{4.282} & \textbf{9.954} & \textbf{8.424} & \textbf{7.864} & \textbf{5.850} & \textbf{9.846} & \textbf{6.828} & \textbf{6.124} & \textbf{7.396} \\ \bottomrule[1.5pt]
\end{tabular}%
}
\vspace{-0.4cm}
\end{table*}

\begin{table*}[t]
\centering
\tabcolsep=0.03cm
\caption{Quantitative comparisons on the MVP dataset for
shape completion (16,384 points) using $\ell1$ chamfer distance $\times 10^{3}$. Lower is better.}
\label{tab:MVP}
\resizebox{\textwidth}{!}{%
\begin{tabular}{c|ccccccccccccccccc}
\toprule[1.5pt]
CD-$\ell1$($\times 10^{3}$) & Chair & Table & Sofa & Cabinet & Lamp & Car & Airplane & Watercraft & Bed & Bench & Bookshelf & Bus & Guitar & Motorbike & Pistol & Skateboard & Avg. \\ \midrule[1pt]
\rowcolor[HTML]{EFEFEF} 
ASFM~\citep{xia2021asfm} & 13.799 & 12.687 & 13.210 & 12.285 & 14.510 & 10.861 & 6.530 & 11.351 & 16.810 & 10.496 & 14.105 & 9.240 & 5.571 & 9.868 & 8.456 & 7.263 & 11.484 \\
CRN~\citep{wang2021cascaded} & 11.367 & 10.389 & 10.825 & 10.064 & 12.111 & 8.641 & 5.495 & 9.186 & 18.480 & 11.440 & 12.509 & 9.507 & 12.745 & 11.346 & 16.773 & 11.478 & 10.579 \\
\rowcolor[HTML]{EFEFEF} 
ECG~\citep{pan2020ecg} & 10.168 & 9.032 & 10.002 & 9.635 & 10.711 & 8.180 & 6.189 & 8.450 & 11.956 & 7.850 & 9.870 & 7.055 & 6.734 & 7.448 & 6.552 & 5.542 & 8.753 \\
FoldingNet~\citep{yang2018foldingnet} & 14.668 & 11.966 & 13.041 & 11.246 & 17.878 & 10.180 & 8.144 & 11.923 & 15.940 & 10.948 & 12.359 & 8.650 & 6.255 & 10.928 & 8.576 & 9.418 & 11.881 \\
\rowcolor[HTML]{EFEFEF} 
GRNet~\citep{xie2020grnet} & 14.705 & 13.149 & 13.531 & 12.202 & 15.735 & 10.466 & 7.134 & 10.833 & 16.535 & 11.252 & 14.026 & 9.769 & 6.684 & 9.492 & 8.396 & 8.719 & 11.817 \\
PCN~\citep{yuan2018pcn} & 17.882 & 14.841 & 14.694 & 11.997 & 22.093 & 10.358 & 7.951 & 13.921 & 20.999 & 14.019 & 15.830 & 8.920 & 5.643 & 10.963 & 9.885 & 7.676 & 13.598 \\
\rowcolor[HTML]{EFEFEF} 
TopNet~\citep{tchapmi2019topnet} & 14.895 & 12.929 & 13.988 & 12.617 & 16.765 & 11.053 & 7.979 & 12.537 & 16.529 & 11.043 & 14.796 & 9.797 & 6.760 & 10.868 & 9.492 & 7.875 & 12.357 \\
PoinTr~\citep{yu2021pointr} & 9.369 & 8.469 & 9.490 & 9.343 & 9.465 & 8.293 & 4.806 & 7.533 & 11.921 & 6.948 & 9.438 & 6.781 & 4.089 & 7.447 & 6.176 & 5.141 & 8.070 \\
\rowcolor[HTML]{EFEFEF}
SnowflakeNet~\citep{xiang2021snowflakenet} & 8.772 & 7.551 & 9.035 & 9.056 & 9.159 & 8.052 & 4.393 & 7.404 & 10.750 & 6.431 & 8.383 & 6.735 & 3.454 & 6.905 & 5.722 & 4.458 & 7.597 \\
\midrule[1pt]
MMPT & \textbf{7.572} & \textbf{6.880} & \textbf{8.131} & \textbf{8.932} & \textbf{6.095} & \textbf{8.048} & \textbf{3.977} & \textbf{6.318} & \textbf{8.874} & \textbf{5.734} & \textbf{7.770} & \textbf{6.581} & \textbf{3.355} & \textbf{6.640} & \textbf{5.367} & \textbf{4.418} & \textbf{6.769} \\ \bottomrule[1.5pt]
\end{tabular}%
}
\vspace{-0.4cm}
\end{table*}

\textbf{Overall Objective.}
In our MMPT, we effectively optimize our model with a joint training loss $\mathcal{L}_{joint}$, consisting of four parts: Reconstruction term $\mathcal{L}_{rec}$, MoCo term $\mathcal{L}_{MoCo}$, intra-modal learning term $\mathcal{L}_{iml}$ and cross-modal learning term $\mathcal{L}_{cml}$. Overall, the joint loss for MMPT pre-training is formulated as:
\begin{equation}
\begin{aligned}
\mathcal{L}_{joint} & = \alpha\mathcal{L}_{rec}+\beta\mathcal{L}_{MoCo}+\gamma\mathcal{L}_{IMID}+\gamma\mathcal{L}_{CMID}
\end{aligned}
\end{equation}
where $\alpha,\beta, \text{and} \gamma$ are hyper-parameters to balance different loss terms.

\textbf{Reconstruction Term.} As an objective of reconstruction learning, we minimize relative distance between predicted patches and ground truth ones through $L_2$-normalized Chamfer Distance (CD) and binary focal loss, which is shown as follows:
\begin{equation}\small
\begin{aligned}
\mathcal{L}_{rec} &= \mathcal{L}_{rec\_cd}(P_{pre},P_{gt}) + \mathcal{L}_{rec\_bce}(Q_{pre},Q_{labels})\\&=\frac{1}{|P_{pre}|}\sum_{p\in P_{pre}}\min_{g\in P_{gt}}\|p-g\|_{2}^{2}+\frac{1}{|P_{gt}|}\sum_{g\in P_{gt}}\min_{p\in P_{pre}}\|g-p\|_{2}^{2}\\&+-\frac{1}{N} {\textstyle \sum_{i}^{N}} \left[l\times \log(p)+(1-l)\times \log(1-p)\right]
\end{aligned}
\end{equation}
where the former of subscripts $p$ and $g$ indicate point patches $P_{pre}$ and $P_{gt}$, while the latter represents point patches in $P_{gt}$ and $P_{pre}$.

\textbf{MoCo Term.} 
Based on our masked point groups prediction module, we can obtain the 3D logits ${S}^{3D}$ for point clouds. The MoCo term $\mathcal{L}_{MoCo}$ mainly applies to help the transformers to better learn the high-level semantic representation. Formally, the MoCo loss $\mathcal{L}_{MoCo}$ can be formulated as: 
\begin{equation}
\begin{aligned}
\mathcal{L}_{MoCo} = \frac{1}{N}\sum_{i=1}^{N}-\log\frac{\exp\left(\mathbf{s}_i^{3D}\cdot\mathbf{s}_{ki}^{labels\prime}/\tau\right)}{\sum_{j=0}^{K}\exp\left({s}_i^{3D}\cdot\mathbf{s}_{kj}^{labels}/\tau\right)}
\end{aligned}
\end{equation}
where ${S}^{3D}$ and ${S}^{labels}$ are the 3D logits and labels, respectively.

\textbf{Intra-Modal Learning Term.} There are two key ideas for loss functions in contrastive learning space that we can employ to train our network. The first idea is the intra-modal learning term, which effectively significantly enhances representation learning for 3D point clouds. Extending Eq~\ref{eq:imid}, we can define our intra-modal learning loss function as:
\begin{equation}
\begin{aligned}
\mathcal{L}_{iml}=\dfrac{1}{2N}\sum_{i=1}^{N}(\ell_{i,{z}^{3D},{s}^{3D}}+\ell_{i,{s}^{3D},{z}^{3D}})
\end{aligned}
\end{equation}

\textbf{Cross-Modal Learning Term.} The second idea is cross-modal learning term between 2D and 3D projected vectors, which improves 2D-3D geometric alignment. Extending Eq~\ref{eq:cmid}, we finally obtain our cross-modal contrastive learning objective $\mathcal{L}_{cml}$ as a combination of $\ell_{i,{s}^{3D},{z}^{2D}}$ and $\ell_{i,{z}^{2D},{s}^{3D}}$:
\begin{equation}
\begin{aligned}
\mathcal{L}_{cmil}=\dfrac{1}{2N}\sum_{i=1}^{N}(\ell_{i,{s}^{3D},{z}^{2D}}+\ell_{i,{z}^{2D},{s}^{3D}})
\end{aligned}
\end{equation}


\begin{figure}[t]
    \centering
    \includegraphics[width=\linewidth]{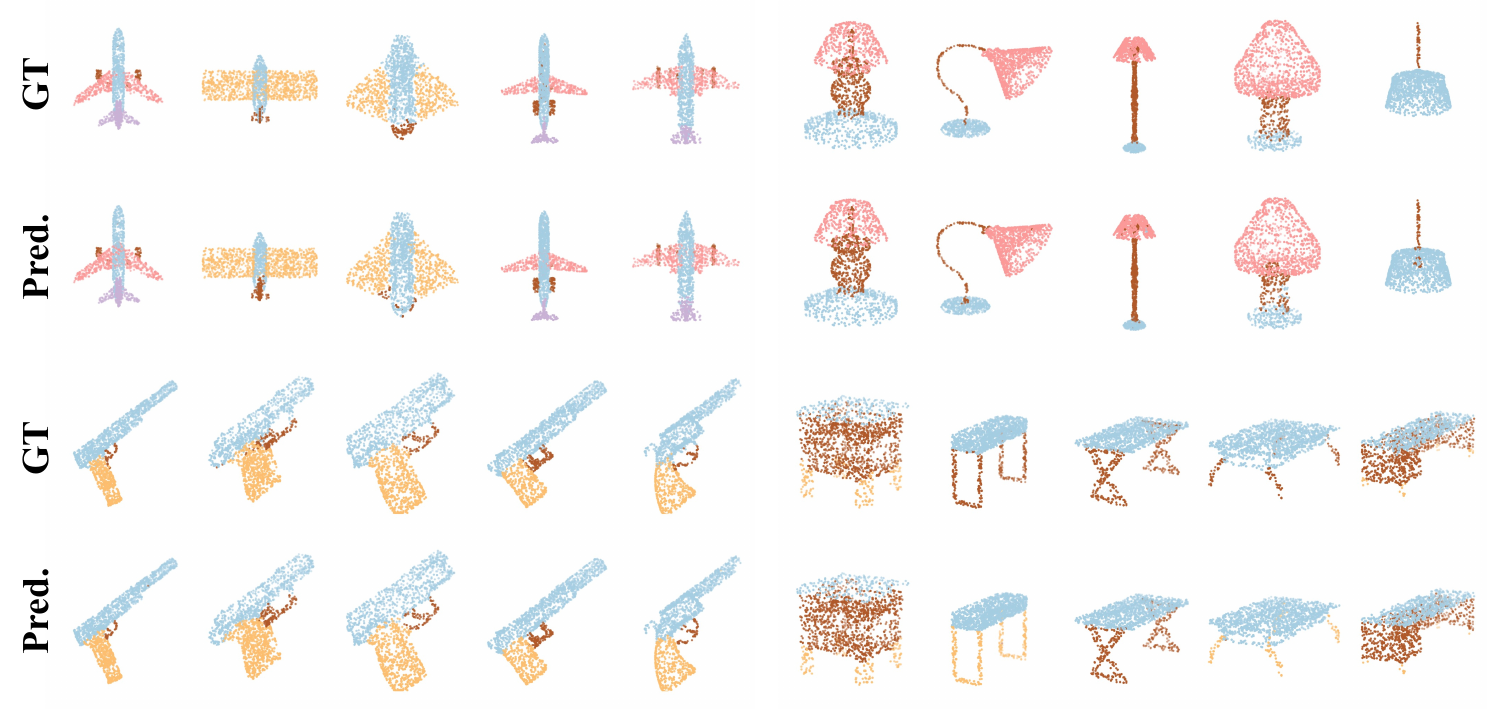}
    \vspace{-0.8cm}
    \caption{Visualization comparison of semantic segmentation on ShapeNetPart dataset by different methods.}
    \label{fig:seg_vis}
\vspace{-0.4cm}
\end{figure}

\begin{figure*}[t]
    \centering
    \includegraphics[width=\linewidth]{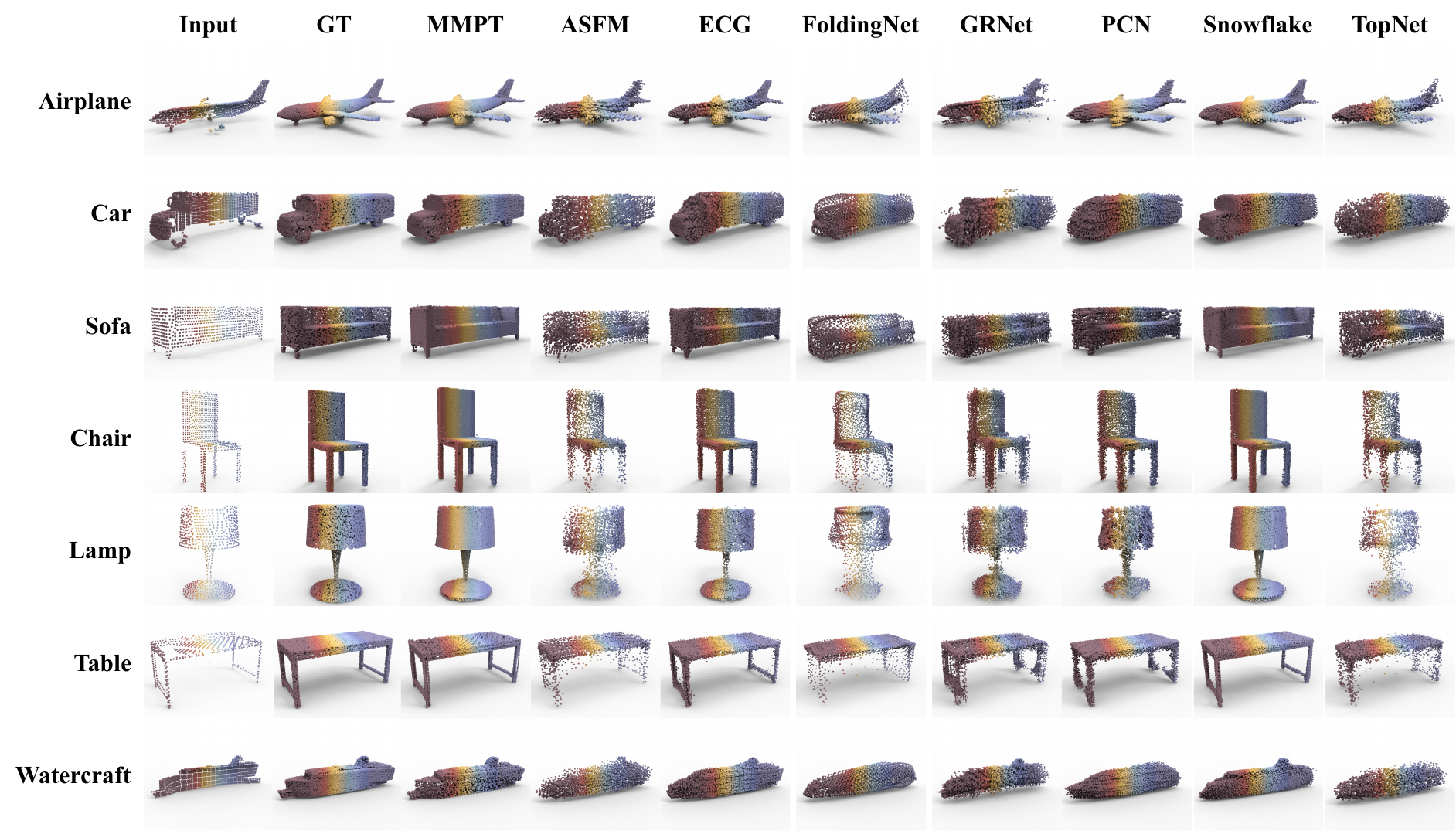}
    \caption{Visualization comparison of point cloud completion on PCN dataset by different methods. From left to right: Point cloud completion based on our MMPT, ASFM~\citep{xia2021asfm}, ECG~\citep{pan2020ecg}, FoldingNet~\citep{yang2018foldingnet}, GRNet~\citep{xie2020grnet}, PCN~\citep{yuan2018pcn}, Snowflake~\citep{xiang2021snowflakenet}, TopNet~\citep{tchapmi2019topnet}, and Ground Truth.}
    \label{fig:PCN}
\end{figure*}
\begin{figure*}[t]
    \centering
    \includegraphics[width=\linewidth]{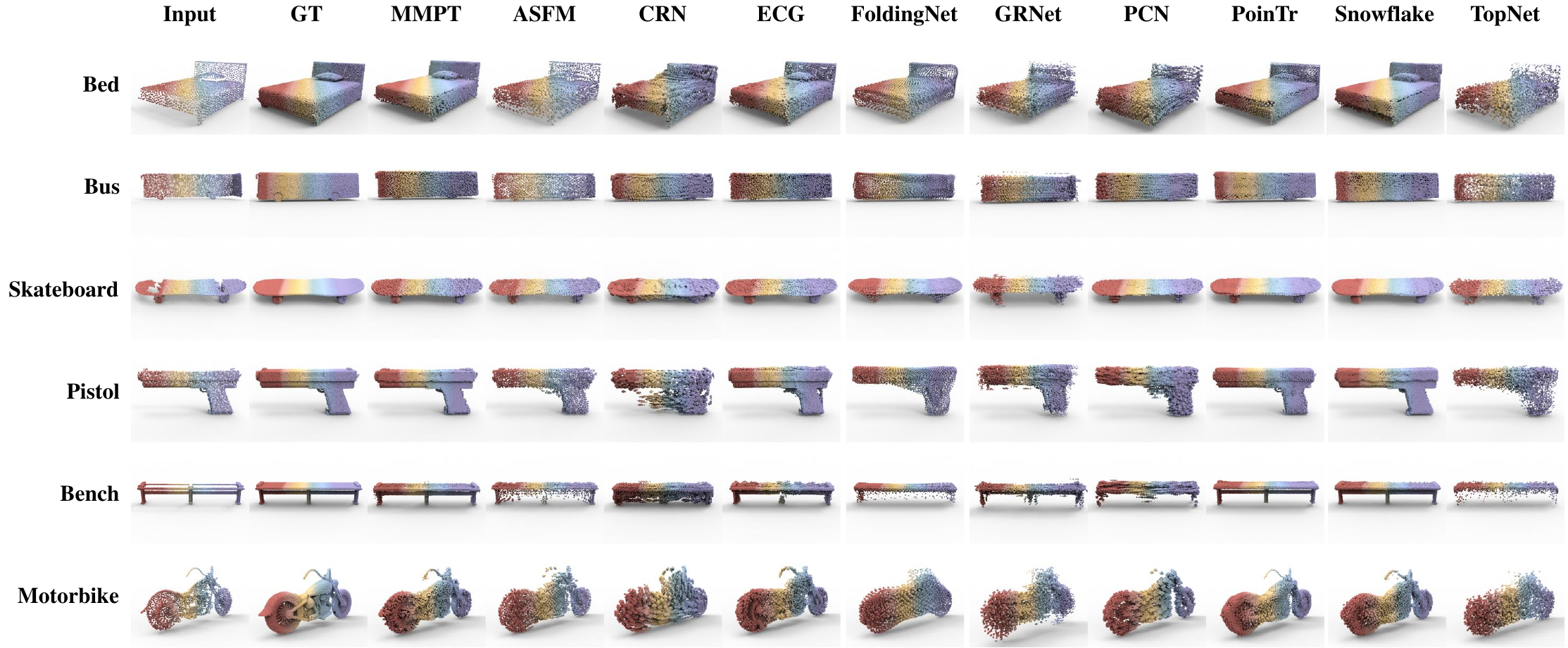}
    \caption{Visualization of qualitative comparison results on the MVP dataset by different methods. From top to bottom: six object categories were randomly selected from sixteen object categories of the MVP dataset.}
    \label{fig:MVP}
\end{figure*}
\begin{figure*}[t]
    \centering
    \subfigure[ShapeNet55]{
    \includegraphics[width=0.48\linewidth, height=0.26\textwidth]{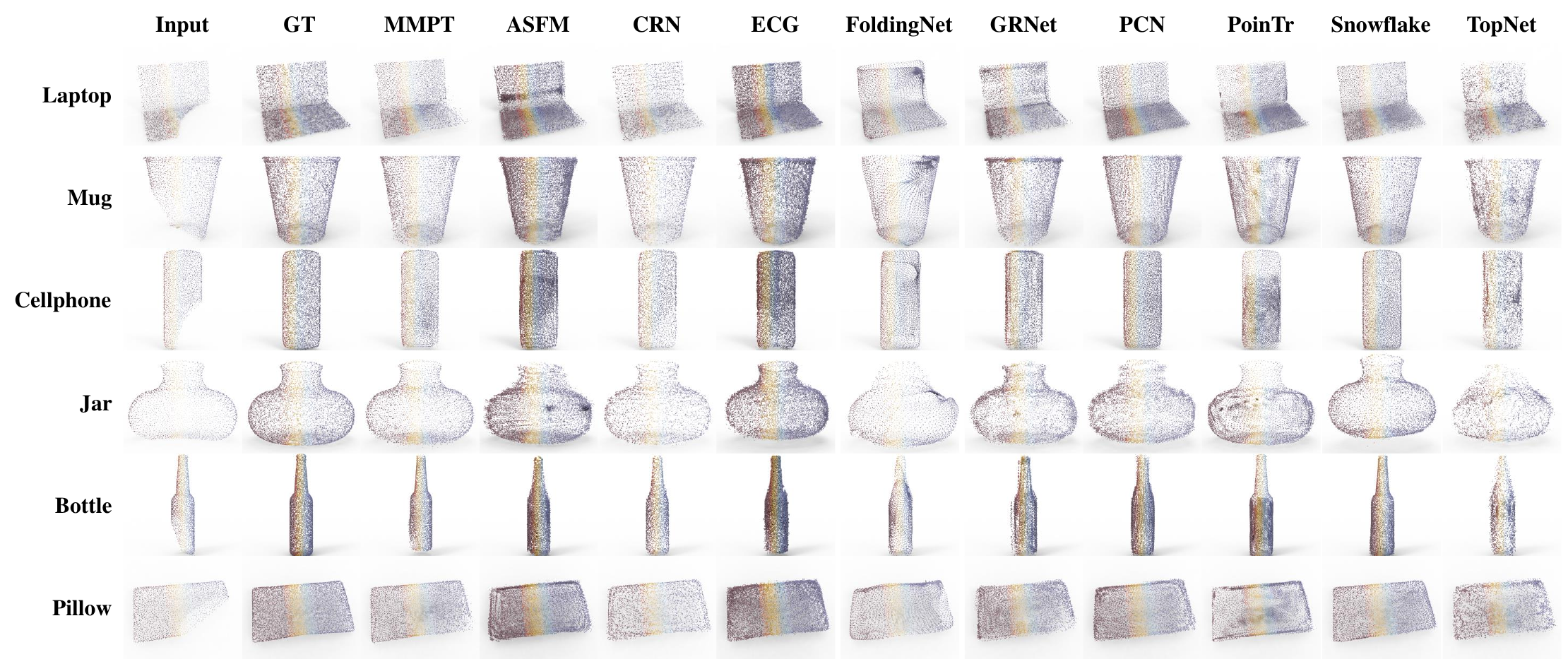}}
    \subfigure[ShapeNetUnseen21]{
    \includegraphics[width=0.48\linewidth, height=0.26\textwidth]{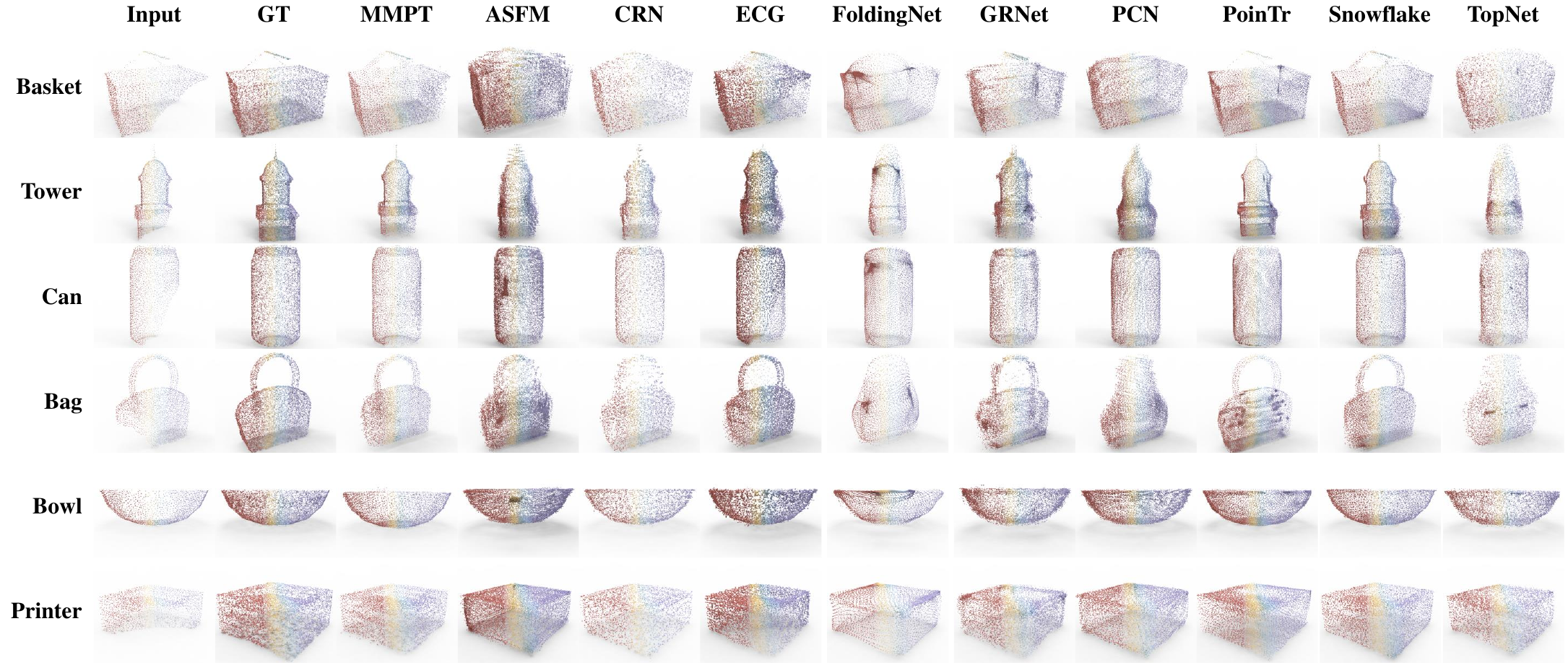}
    }
    \caption{Visualization of qualitative comparison results on the ShapeNet55/21 dataset by different methods. From left to right: point cloud completion based on our MMPT, ASFM, CRN, ECG, FoldingNet, GRNet, PCN, PoinTr, Snowflake, and TopNet.}
    \label{fig:ShapeNet55/21}
\end{figure*}

\begin{figure}[t]
    \centering
    \includegraphics[width=\linewidth]{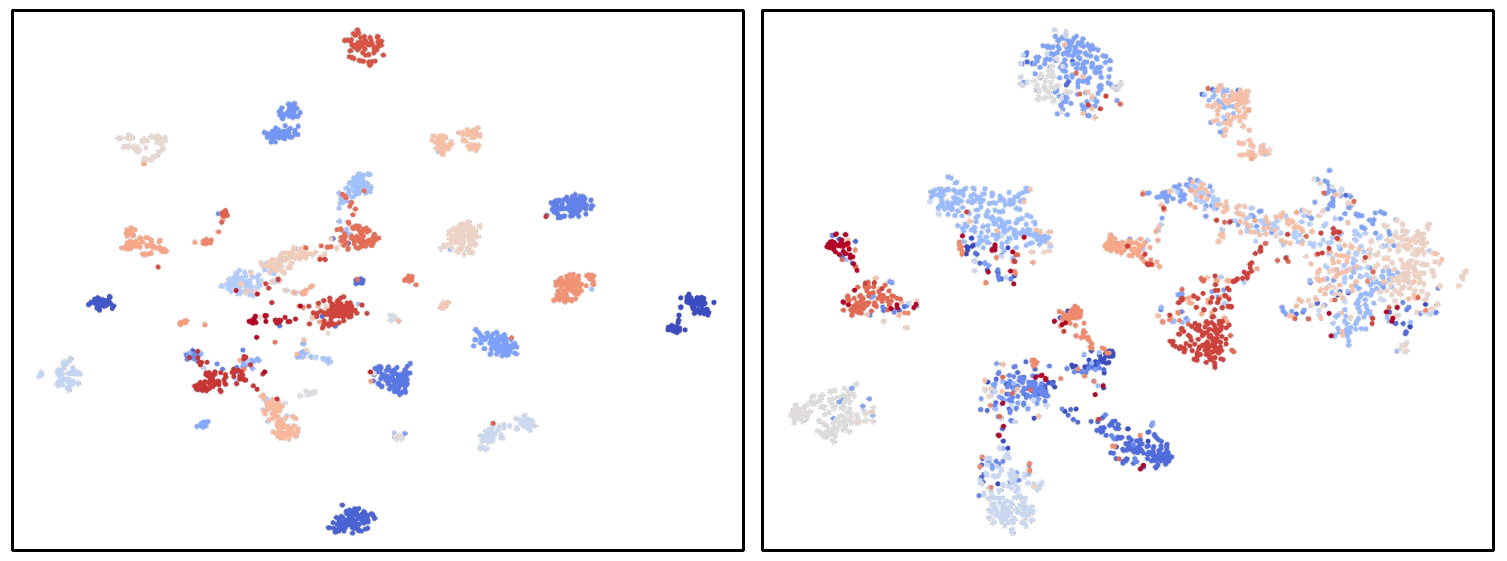}
    \caption{t-SNE visualization of the features learned from ModelNet40 and ScanObjectNN (OBJ-ONLY).}
    \label{fig:tsne}
\end{figure}

\begin{table*}[t]
\centering
\tabcolsep=0.03cm
\caption{The performance comparison of MMPT and other networks on ShapeNet55 in terms of the average CD-$\ell1$ $\times 10^{3}$, CD-$\ell2$ $\times 10^{3}$ and F-Score$@1\%$. Three difficulties CD-S, CD-M, and CD-H are employed to measure the completion results, which represent the Simple, Moderate, and Hard settings.}
\label{tab:ShapeNet55}
\resizebox{\textwidth}{!}{%
\begin{tabular}{c|ccccc|cccccccccc}
\toprule[1.5pt]
 &  &  &  &  &  &  &  &  &  &  &  &  &  &  &  \\
 &  &  &  &  &  &  &  &  &  &  &  &  &  &  &  \\
 &  &  &  &  &  &  &  &  &  &  &  &  &  &  &  \\
\multirow{-4}{*}{\textbf{}} & \multirow{-4}{*}{F1-Avg} & \multirow{-4}{*}{\begin{tabular}[c]{@{}c@{}}CD-S\\ (CD-$\ell1$/\\ CD-$\ell2$)\end{tabular}} & \multirow{-4}{*}{\begin{tabular}[c]{@{}c@{}}CD-M\\ (CD-$\ell1$/\\ CD-$\ell2$)\end{tabular}} & \multirow{-4}{*}{\begin{tabular}[c]{@{}c@{}}CD-H\\ (CD-$\ell1$/\\ CD-$\ell2$)\end{tabular}} & \multirow{-4}{*}{\begin{tabular}[c]{@{}c@{}}CD-Avg\\ (CD-$\ell1$/\\ CD-$\ell2$)\end{tabular}} & \multirow{-4}{*}{\begin{tabular}[c]{@{}c@{}}Table\\ (F-Score-Avg\\ /CD-$\ell1$-Avg\\ /CD-$\ell2$-Avg)\end{tabular}} & \multirow{-4}{*}{\begin{tabular}[c]{@{}c@{}}Chair\\ (F-Score-Avg\\ /CD-$\ell1$-Avg\\ /CD-$\ell2$-Avg)\end{tabular}} & \multirow{-4}{*}{\begin{tabular}[c]{@{}c@{}}Airplane\\ (F-Score-Avg\\ /CD-$\ell1$-Avg\\ /CD-$\ell2$-Avg)\end{tabular}} & \multirow{-4}{*}{\begin{tabular}[c]{@{}c@{}}Car\\ (F-Score-Avg\\ /CD-$\ell1$-Avg\\ /CD-$\ell2$-Avg)\end{tabular}} & \multirow{-4}{*}{\begin{tabular}[c]{@{}c@{}}Sofa\\ (F-Score-Avg\\ /CD-$\ell1$-Avg\\ /CD-$\ell2$-Avg)\end{tabular}} & \multirow{-4}{*}{\begin{tabular}[c]{@{}c@{}}Birdhouse\\ (F-Score-Avg\\ /CD-$\ell1$-Avg\\ /CD-$\ell2$-Avg)\end{tabular}} & \multirow{-4}{*}{\begin{tabular}[c]{@{}c@{}}Bag\\ (F-Score-Avg\\ /CD-$\ell1$-Avg\\ /CD-$\ell2$-Avg)\end{tabular}} & \multirow{-4}{*}{\begin{tabular}[c]{@{}c@{}}Remote\\ (F-Score-Avg\\ /CD-$\ell1$-Avg\\ /CD-$\ell2$-Avg)\end{tabular}} & \multirow{-4}{*}{\begin{tabular}[c]{@{}c@{}}Keyboard\\ (F-Score-Avg\\ /CD-$\ell1$-Avg\\ /CD-$\ell2$-Avg)\end{tabular}} & \multirow{-4}{*}{\begin{tabular}[c]{@{}c@{}}Rocket\\ (F-Score-Avg\\ /CD-$\ell1$-Avg\\ /CD-$\ell2$-Avg)\end{tabular}} \\ \midrule[1pt]
\rowcolor[HTML]{EFEFEF} 
\cellcolor[HTML]{EFEFEF} & \cellcolor[HTML]{EFEFEF} & \cellcolor[HTML]{EFEFEF} & \cellcolor[HTML]{EFEFEF} & \cellcolor[HTML]{EFEFEF} & \cellcolor[HTML]{EFEFEF} & 0.294 & 0.209 & 0.426 & 0.164 & 0.177 & 0.122 & 0.219 & 0.332 & 0.357 & 0.522 \\
\rowcolor[HTML]{EFEFEF} 
\cellcolor[HTML]{EFEFEF} & \cellcolor[HTML]{EFEFEF} & \cellcolor[HTML]{EFEFEF} & \cellcolor[HTML]{EFEFEF} & \cellcolor[HTML]{EFEFEF} & \cellcolor[HTML]{EFEFEF} & 17.790 & 21.217 & 14.244 & 21.077 & 20.876 & 27.943 & 21.097 & 16.142 & 14.258 & 13.892 \\
\rowcolor[HTML]{EFEFEF} 
\multirow{-3}{*}{\cellcolor[HTML]{EFEFEF}ASFM~\citep{xia2021asfm}} & \multirow{-3}{*}{\cellcolor[HTML]{EFEFEF}0.247} & \multirow{-3}{*}{\cellcolor[HTML]{EFEFEF}\begin{tabular}[c]{@{}c@{}}19.138\\ 1.308\end{tabular}} & \multirow{-3}{*}{\cellcolor[HTML]{EFEFEF}\begin{tabular}[c]{@{}c@{}}20.172\\ 1.517\end{tabular}} & \multirow{-3}{*}{\cellcolor[HTML]{EFEFEF}\begin{tabular}[c]{@{}c@{}}23.513\\ 2.282\end{tabular}} & \multirow{-3}{*}{\cellcolor[HTML]{EFEFEF}\begin{tabular}[c]{@{}c@{}}20.941\\ 1.702\end{tabular}} & 1.241 & 1.594 & 0.749 & 1.293 & 1.371 & 2.769 & 1.620 & 0.882 & 0.677 & 0.866 \\
 &  &  &  &  &  & 0.214 & 0.183 & 0.357 & 0.118 & 0.142 & 0.102 & 0.163 & 0.247 & 0.266 & 0.485 \\
 &  &  &  &  &  & 20.313 & 22.931 & 15.231 & 23.804 & 23.367 & 32.237 & 24.139 & 18.164 & 15.947 & 14.522 \\
\multirow{-3}{*}{CRN~\citep{wang2021cascaded}} & \multirow{-3}{*}{0.205} & \multirow{-3}{*}{\begin{tabular}[c]{@{}c@{}}21.207\\ 1.502\end{tabular}} & \multirow{-3}{*}{\begin{tabular}[c]{@{}c@{}}22.364\\ 1.801\end{tabular}} & \multirow{-3}{*}{\begin{tabular}[c]{@{}c@{}}25.849\\ 2.726\end{tabular}} & \multirow{-3}{*}{\begin{tabular}[c]{@{}c@{}}23.140\\ 2.010\end{tabular}} & 1.570 & 1.859 & 0.842 & 1.606 & 1.663 & 3.599 & 2.040 & 1.041 & 0.815 & 0.930 \\
\rowcolor[HTML]{EFEFEF} 
\cellcolor[HTML]{EFEFEF} & \cellcolor[HTML]{EFEFEF} & \cellcolor[HTML]{EFEFEF} & \cellcolor[HTML]{EFEFEF} & \cellcolor[HTML]{EFEFEF} & \cellcolor[HTML]{EFEFEF} & 0.320 & 0.336 & 0.558 & 0.254 & 0.234 & 0.194 & 0.284 & 0.374 & 0.454 & 0.639 \\
\rowcolor[HTML]{EFEFEF} 
\cellcolor[HTML]{EFEFEF} & \cellcolor[HTML]{EFEFEF} & \cellcolor[HTML]{EFEFEF} & \cellcolor[HTML]{EFEFEF} & \cellcolor[HTML]{EFEFEF} & \cellcolor[HTML]{EFEFEF} & 17.914 & 18.234 & 12.349 & 18.163 & 19.815 & 25.934 & 20.013 & 15.868 & 12.890 & 11.745 \\
\rowcolor[HTML]{EFEFEF} 
\multirow{-3}{*}{\cellcolor[HTML]{EFEFEF}ECG~\citep{pan2020ecg}} & \multirow{-3}{*}{\cellcolor[HTML]{EFEFEF}0.321} & \multirow{-3}{*}{\cellcolor[HTML]{EFEFEF}\begin{tabular}[c]{@{}c@{}}16.710\\ 1.167\end{tabular}} & \multirow{-3}{*}{\cellcolor[HTML]{EFEFEF}\begin{tabular}[c]{@{}c@{}}18.727\\ 1.545\end{tabular}} & \multirow{-3}{*}{\cellcolor[HTML]{EFEFEF}\begin{tabular}[c]{@{}c@{}}23.480\\ 2.555\end{tabular}} & \multirow{-3}{*}{\cellcolor[HTML]{EFEFEF}\begin{tabular}[c]{@{}c@{}}19.639\\ 1.756\end{tabular}} & 1.543 & 1.409 & 0.657 & 1.034 & 1.422 & 2.832 & 1.676 & 0.961 & 0.688 & 0.702 \\
 &  &  &  &  &  & 0.163 & 0.066 & 0.172 & 0.063 & 0.061 & 0.027 & 0.066 & 0.139 & 0.201 & 0.209 \\
 &  &  &  &  &  & 23.298 & 27.194 & 19.750 & 25.383 & 26.752 & 36.925 & 27.559 & 20.136 & 18.123 & 19.424 \\
\multirow{-3}{*}{FoldingNet~\citep{yang2018foldingnet}} & \multirow{-3}{*}{0.091} & \multirow{-3}{*}{\begin{tabular}[c]{@{}c@{}}25.203\\ 2.095\end{tabular}} & \multirow{-3}{*}{\begin{tabular}[c]{@{}c@{}}26.596\\ 2.410\end{tabular}} & \multirow{-3}{*}{\begin{tabular}[c]{@{}c@{}}30.424\\ 3.333\end{tabular}} & \multirow{-3}{*}{\begin{tabular}[c]{@{}c@{}}27.408\\ 2.613\end{tabular}} & 2.043 & 2.440 & 1.331 & 1.777 & 2.076 & 4.291 & 2.439 & 1.209 & 1.054 & 1.260 \\
\rowcolor[HTML]{EFEFEF} 
\cellcolor[HTML]{EFEFEF} & \cellcolor[HTML]{EFEFEF} & \cellcolor[HTML]{EFEFEF} & \cellcolor[HTML]{EFEFEF} & \cellcolor[HTML]{EFEFEF} & \cellcolor[HTML]{EFEFEF} & 0.231 & 0.221 & 0.412 & 0.150 & 0.168 & 0.136 & 0.210 & 0.307 & 0.294 & 0.535 \\
\rowcolor[HTML]{EFEFEF} 
\cellcolor[HTML]{EFEFEF} & \cellcolor[HTML]{EFEFEF} & \cellcolor[HTML]{EFEFEF} & \cellcolor[HTML]{EFEFEF} & \cellcolor[HTML]{EFEFEF} & \cellcolor[HTML]{EFEFEF} & 19.397 & 21.213 & 14.829 & 21.927 & 22.024 & 27.332 & 22.041 & 17.273 & 15.661 & 13.602 \\
\rowcolor[HTML]{EFEFEF} 
\multirow{-3}{*}{\cellcolor[HTML]{EFEFEF}GRNet~\citep{xie2020grnet}} & \multirow{-3}{*}{\cellcolor[HTML]{EFEFEF}0.239} & \multirow{-3}{*}{\cellcolor[HTML]{EFEFEF}\begin{tabular}[c]{@{}c@{}}19.159\\ 1.137\end{tabular}} & \multirow{-3}{*}{\cellcolor[HTML]{EFEFEF}\begin{tabular}[c]{@{}c@{}}20.645\\ 1.489\end{tabular}} & \multirow{-3}{*}{\cellcolor[HTML]{EFEFEF}\begin{tabular}[c]{@{}c@{}}24.034\\ 2.394\end{tabular}} & \multirow{-3}{*}{\cellcolor[HTML]{EFEFEF}\begin{tabular}[c]{@{}c@{}}21.279\\ 1.673\end{tabular}} & 1.354 & 1.575 & 0.830 & 1.351 & 1.435 & 2.579 & 1.700 & 0.954 & 0.780 & 0.855 \\
 &  &  &  &  &  & 0.181 & 0.143 & 0.340 & 0.103 & 0.101 & 0.062 & 0.127 & 0.228 & 0.272 & 0.343 \\
 &  &  &  &  &  & 21.695 & 24.558 & 16.428 & 23.398 & 24.364 & 34.575 & 26.025 & 18.419 & 16.272 & 18.587 \\
\multirow{-3}{*}{PCN~\citep{yuan2018pcn}} & \multirow{-3}{*}{0.167} & \multirow{-3}{*}{\begin{tabular}[c]{@{}c@{}}22.990\\ 1.811\end{tabular}} & \multirow{-3}{*}{\begin{tabular}[c]{@{}c@{}}23.976\\ 2.062\end{tabular}} & \multirow{-3}{*}{\begin{tabular}[c]{@{}c@{}}27.360\\ 2.937\end{tabular}} & \multirow{-3}{*}{\begin{tabular}[c]{@{}c@{}}24.775\\ 2.270\end{tabular}} & 1.812 & 2.085 & 0.985 & 1.532 & 1.765 & 4.023 & 2.294 & 1.041 & 0.924 & 1.318 \\
\rowcolor[HTML]{EFEFEF} 
\cellcolor[HTML]{EFEFEF} & \cellcolor[HTML]{EFEFEF} & \cellcolor[HTML]{EFEFEF} & \cellcolor[HTML]{EFEFEF} & \cellcolor[HTML]{EFEFEF} & \cellcolor[HTML]{EFEFEF} & 0.147 & 0.088 & 0.203 & 0.077 & 0.077 & 0.046 & 0.084 & 0.121 & 0.218 & 0.278 \\
\rowcolor[HTML]{EFEFEF} 
\cellcolor[HTML]{EFEFEF} & \cellcolor[HTML]{EFEFEF} & \cellcolor[HTML]{EFEFEF} & \cellcolor[HTML]{EFEFEF} & \cellcolor[HTML]{EFEFEF} & \cellcolor[HTML]{EFEFEF} & 25.106 & 29.754 & 20.034 & 27.212 & 28.739 & 38.762 & 30.489 & 25.026 & 18.207 & 19.874 \\
\rowcolor[HTML]{EFEFEF} 
\multirow{-3}{*}{\cellcolor[HTML]{EFEFEF}TopNet~\citep{tchapmi2019topnet}} & \multirow{-3}{*}{\cellcolor[HTML]{EFEFEF}0.110} & \multirow{-3}{*}{\cellcolor[HTML]{EFEFEF}\begin{tabular}[c]{@{}c@{}}27.233\\ 2.483\end{tabular}} & \multirow{-3}{*}{\cellcolor[HTML]{EFEFEF}\begin{tabular}[c]{@{}c@{}}28.749\\ 2.848\end{tabular}} & \multirow{-3}{*}{\cellcolor[HTML]{EFEFEF}\begin{tabular}[c]{@{}c@{}}33.986\\ 4.642\end{tabular}} & \multirow{-3}{*}{\cellcolor[HTML]{EFEFEF}\begin{tabular}[c]{@{}c@{}}29.989\\ 3.324\end{tabular}} & 2.438 & 2.950 & 1.397 & 2.089 & 2.449 & 4.899 & 2.998 & 1.982 & 1.091 & 1.498 \\
 &  &  &  &  &  & \textbf{0.480} & \textbf{0.438} & \textbf{0.598} & \textbf{0.368} & \textbf{0.394} & \textbf{0.348} & \textbf{0.416} & \textbf{0.510} & 0.533 & \textbf{0.690} \\
 &  &  &  &  &  & 13.041 & 14.838 & 10.137 & 15.747 & 14.790 & 20.672 & 15.297 & 11.660 & 9.709 & 9.841 \\
\multirow{-3}{*}{PoinTr~\citep{yu2021pointr}} & \multirow{-3}{*}{\textbf{0.446}} & \multirow{-3}{*}{\begin{tabular}[c]{@{}c@{}}12.491\\ 0.698\end{tabular}} & \multirow{-3}{*}{\begin{tabular}[c]{@{}c@{}}14.182\\ 1.049\end{tabular}} & \multirow{-3}{*}{\begin{tabular}[c]{@{}c@{}}18.811\\ 2.022\end{tabular}} & \multirow{-3}{*}{\begin{tabular}[c]{@{}c@{}}15.161\\ 1.256\end{tabular}} & 0.979 & 1.149 & 0.547 & 0.974 & 0.944 & 2.132 & 1.170 & 0.621 & \textbf{0.400} & 0.627 \\
\rowcolor[HTML]{EFEFEF} 
\cellcolor[HTML]{EFEFEF} & \cellcolor[HTML]{EFEFEF} & \cellcolor[HTML]{EFEFEF} & \cellcolor[HTML]{EFEFEF} & \cellcolor[HTML]{EFEFEF} & \cellcolor[HTML]{EFEFEF} & 0.379 & 0.362 & 0.540 & 0.232 & 0.297 & 0.244 & 0.347 & 0.450 & 0.452 & 0.652 \\
\rowcolor[HTML]{EFEFEF} 
\cellcolor[HTML]{EFEFEF} & \cellcolor[HTML]{EFEFEF} & \cellcolor[HTML]{EFEFEF} & \cellcolor[HTML]{EFEFEF} & \cellcolor[HTML]{EFEFEF} & \cellcolor[HTML]{EFEFEF} & 14.275 & 15.706 & 11.091 & 17.443 & 16.177 & 21.162 & 16.121 & 12.457 & 11.263 & 10.528 \\
\rowcolor[HTML]{EFEFEF} 
\multirow{-3}{*}{\cellcolor[HTML]{EFEFEF}SnowflakeNet~\citep{xiang2021snowflakenet}} & \multirow{-3}{*}{\cellcolor[HTML]{EFEFEF}0.362} & \multirow{-3}{*}{\cellcolor[HTML]{EFEFEF}\begin{tabular}[c]{@{}c@{}}13.568\\ 0.680\end{tabular}} & \multirow{-3}{*}{\cellcolor[HTML]{EFEFEF}\begin{tabular}[c]{@{}c@{}}15.380\\ \textbf{0.979}\end{tabular}} & \multirow{-3}{*}{\cellcolor[HTML]{EFEFEF}\begin{tabular}[c]{@{}c@{}}19.412\\ \textbf{1.754}\end{tabular}} & \multirow{-3}{*}{\cellcolor[HTML]{EFEFEF}\begin{tabular}[c]{@{}c@{}}16.120\\ \textbf{1.138}\end{tabular}} & \textbf{0.880} & \textbf{1.049} & \textbf{0.529} & \textbf{1.010} & \textbf{0.916} & 1.881 & 1.078 & 0.581 & 0.450 & 0.605 \\ \midrule[1pt]
 &  &  &  &  &  & 0.451 & 0.414 & 0.563 & 0.269 & 0.378 & 0.306 & 0.402 & 0.495 & \textbf{0.538} & 0.686 \\
 &  &  &  &  &  & \textbf{11.885} & \textbf{12.813} & \textbf{9.227} & \textbf{14.536} & \textbf{12.883} & \textbf{16.483} & \textbf{12.612} & \textbf{9.746} & \textbf{8.752} & \textbf{8.040} \\
\multirow{-3}{*}{MMPT} & \multirow{-3}{*}{0.410} & \multirow{-3}{*}{\begin{tabular}[c]{@{}c@{}}\textbf{10.416}\\ \textbf{0.632}\end{tabular}} & \multirow{-3}{*}{\begin{tabular}[c]{@{}c@{}}\textbf{12.455}\\ 1.054\end{tabular}} & \multirow{-3}{*}{\begin{tabular}[c]{@{}c@{}}\textbf{17.093}\\2.157\end{tabular}} & \multirow{-3}{*}{\begin{tabular}[c]{@{}c@{}}\textbf{13.321}\\ 1.281\end{tabular}} & 1.057 & 1.150 & 0.605 & 1.017 & 0.979 & \textbf{1.679} & \textbf{1.029} & \textbf{0.540} & 0.443 & \textbf{0.519} \\ \bottomrule[1.5pt]
\end{tabular}%
}
\end{table*}

\begin{table*}[t]
\centering
\tabcolsep=0.03cm
\caption{The performance comparison of MMPT and other networks on ShapeNet34 and ShapeNetUnseen21 in terms of the average CD-$\ell1$ $\times 10^{3}$, CD-$\ell2$ $\times 10^{3}$ and F-Score$@1\%$.}
\label{tab:ShapeNet34-21}
\resizebox{\textwidth}{!}{%
\begin{tabular}{c|ccccc|ccccc}
\toprule[1.5pt]
 & \multicolumn{5}{c|}{ShapeNet34} & \multicolumn{5}{c}{ShapeNetUnseen21} \\ \midrule[1pt]
 &  &  &  &  &  &  &  &  &  &  \\
\multirow{-2}{*}{\textbf{}} & \multirow{-2}{*}{\begin{tabular}[c]{@{}c@{}}CD-S\\ (CD-$\ell1$/CD-$\ell2$)\end{tabular}} & \multirow{-2}{*}{\begin{tabular}[c]{@{}c@{}}CD-M\\ (CD-$\ell1$/CD-$\ell2$)\end{tabular}} & \multirow{-2}{*}{\begin{tabular}[c]{@{}c@{}}CD-H\\ (CD-$\ell1$/CD-$\ell2$)\end{tabular}} & \multirow{-2}{*}{\begin{tabular}[c]{@{}c@{}}CD-Avg\\ (CD-$\ell1$/CD-$\ell2$)\end{tabular}} & \multirow{-2}{*}{F1-Avg} & \multirow{-2}{*}{\begin{tabular}[c]{@{}c@{}}CD-S\\ (CD-$\ell1$/CD-$\ell2$)\end{tabular}} & \multirow{-2}{*}{\begin{tabular}[c]{@{}c@{}}CD-M\\ (CD-$\ell1$/CD-$\ell2$)\end{tabular}} & \multirow{-2}{*}{\begin{tabular}[c]{@{}c@{}}CD-H\\ (CD-$\ell1$/CD-$\ell2$)\end{tabular}} & \multirow{-2}{*}{\begin{tabular}[c]{@{}c@{}}CD-Avg\\ (CD-$\ell1$/CD-$\ell2$)\end{tabular}} & \multirow{-2}{*}{F1-Avg} \\ \midrule[1pt]
\rowcolor[HTML]{EFEFEF} 
ASFM~\citep{xia2021asfm} & 18.350/1.189 & 19.123/1.343 & 21.913/1.909 & 19.795/1.480 & 0.268 & 21.589/1.995 & 23.006/2.342 & 27.628/3.660 & 24.074/2.666 & 0.216 \\
CRN~\citep{wang2021cascaded} & 20.304/1.362 & 21.216/1.594 & 24.159/2.318 & 21.893/1.758 & 0.221 & 24.247/2.237 & 26.076/2.840 & 31.771/4.833 & 27.365/3.303 & 0.177 \\
\rowcolor[HTML]{EFEFEF} 
ECG~\citep{pan2020ecg} & 13.122/0.735 & 14.628/0.996 & 18.461/1.696 & 15.404/1.142 & \textbf{0.496} & 15.282/1.255 & 17.595/1.759 & 23.535/3.267 & 18.804/2.094 & \textbf{0.460} \\
FoldingNet~\citep{yang2018foldingnet} & 23.556/1.859 & 24.466/2.059 & 27.584/2.759 & 25.202/2.226 & 0.137 & 28.356/2.887 & 29.833/3.290 & 35.356/4.968 & 31.182/3.715 & 0.088 \\
\rowcolor[HTML]{EFEFEF} 
GRNet~\citep{xie2020grnet} & 18.809/1.102 & 20.034/1.366 & 22.989/2.089 & 20.611/1.519 & 0.247 & 21.246/1.553 & 23.753/2.281 & 29.427/4.169 & 24.809/2.668 & 0.208 \\
PCN~\citep{yuan2018pcn} & 21.433/1.551 & 22.304/1.753 & 25.086/2.426 & 22.941/1.910 & 0.192 & 27.593/2.983 & 28.989/3.442 & 34.598/5.558 & 30.393/3.994 & 0.128 \\
\rowcolor[HTML]{EFEFEF} 
TopNet~\citep{tchapmi2019topnet} & 22.382/1.606 & 23.271/1.793 & 26.020/2.432 & 23.891/1.944 & 0.154 & 26.775/2.499 & 28.312/2.928 & 33.121/4.407 & 29.403/3.278 & 0.103 \\
PoinTr~\citep{yu2021pointr} & 12.006/0.632 & 13.393/0.910 & 17.365/\textbf{1.697} & 14.255/1.080 & 0.459 & 13.290/0.838 & 15.521/1.376 & 21.881/3.070 & 16.897/1.761 & 0.421 \\
\rowcolor[HTML]{EFEFEF} 
SnowflakeNet~\citep{xiang2021snowflakenet} & 13.612/0.693 & 15.272/0.968 & 19.385/1.727 & 16.090 /1.129 & 0.370 & 15.162/0.974 & 17.720/1.491 & 23.986/3.022 & 18.956/1.829 & 0.331 \\ \hline
MMPT & \textbf{10.062/0.570} & \textbf{11.717/0.888} & \textbf{15.561}/1.729 & \textbf{12.447/1.062} & 0.429
& \textbf{10.742/0.711} & \textbf{13.090/1.229} & \textbf{18.710/2.709} & \textbf{14.181/1.550} & 0.396\\ \bottomrule[1.5pt]
\end{tabular}%
}
\end{table*}

\section{Experiments}

In this section, we introduce the pre-training setups and the performance of downstream tasks. The details about datasets and fine-tuning setups on downstream tasks can be found in Supplementary Material.

\subsection{Pre-training Setup}

\textbf{Pre-training Datasets.}
We use ShapeNetRender~\citep{afham2022crosspoint} as our pre-training dataset for several downstream point cloud understanding tasks. Additionally, we also utilize colored single-view pictures from the ShapeNetRender~\citep{afham2022crosspoint} dataset. 
Each RGB image is associated with a depth image, a normal map, and an albedo image, with greater variety in the camera angles.

\textbf{Transformer Architecture.}
Our goal is to develop a pre-trained model with robust generalization capabilities by multi-task pre-training. 
We utilize two separate transformers: a Token-Level Transformer Auto-Encoder to obtain the point feature, and the MaskTransformer~\citep{liu2022masked} for point-level reconstruction. Following the inspiration of Point-BERT\citep{yu2022point}, we construct a 12-layer standard transformer encoder in the Token-Level Transformer Auto-Encoder. The hidden dimension of each encoder block is set to 384, with 6 heads, a Feed Forward Network (FFN) expansion ratio of 4, and a stochastic depth drop rate of 0.1.
Note that these two Transformers share the same encoder in our experiments.

\textbf{Pre-training Details.}
Consistent with~\citep{yu2022point}, we pre-train using the AdamW optimizer with a weight decay of 0.05 and a learning rate of 5 $\times 10^{-4}$ that decays cosinusoidally. 
The model is trained with a batch size of 4 for 100 epochs and includes random scaling and translation data augmentation.


\subsection{Downstream Tasks}

\subsubsection{\textbf{3D Object Classification on Synthetic Data.}}
To evaluate our method on the synthetic dataset, we utilized the ModelNet40 benchmark for 3D object classification. As displayed in Table~\ref{table:modelnet40}, the top section of the table presents the results of the fully-supervised methods, including PointNet~\citep{qi2017pointnet}, PointNet++~\citep{qi2017pointnet++}, PointWeb~\citep{zhao2019pointweb}, and others.
The bottom part of the table presents the current state-of-the-art self-supervised methods, including OcCo~\citep{wang2021unsupervised}, STRL~\citep{huang2021spatio}, Transformer-OcCo~\citep{wang2021unsupervised}, Point-BERT~\citep{yu2022point}, and Point-MAE~\citep{pang2022masked}. 
Based on the results, our method achieves a comparative performance of 93.9$\%$ accuracy on ModelNet40, surpassing the performance of comparable methods and setting new benchmark results. Specifically, our method improves accuracy by 0.9$\%$, 0.8$\%$, 1.8$\%$, 0.7$\%$, and 0.8$\%$ compared to OcCo, STRL, Transformer-OcCo, Point-BERT, and Point-MAE, respectively.

\subsubsection{\textbf{3D Object Classification on Real-world Data.}}
To evaluate the effectiveness of our method on real-world data, we utilized the ScanObjectNN benchmark dataset for 3D object classification. We used classification accuracy as the evaluation metric, and the results are presented in Table~\ref{table:scanobjectnn}. To demonstrate the efficacy of our approach, we compared it with both supervised and self-supervised classification methods. 
The results demonstrate that our MMPT method achieves an accuracy of 86.4$\%$ on the most challenging variant PB-T50-RS, which is significantly better than the sophisticated Point-BERT by 3.4$\%$ in accuracy. 
Additionally, comparing the results of ModelNet40 and ScanObjectNN, it is evident that our method achieves remarkable performance on the latter and sets a new state-of-the-art. 
This underscores the significance of utilizing more extensive datasets in our proposed pre-training task to enhance feature representation.

\subsubsection{\textbf{3D Part Segmentation.}}
3D part segmentation is a task that involves predicting the part category label of each point in a point cloud. Our competitors can be broadly categorized into two groups: supervised methods, including PointNet~\citep{qi2017pointnet}, PointNet++~\citep{qi2017pointnet++}, DGCNN~\citep{wang2019dynamic}, and Transformer~\citep{yu2022point}; and self-supervised methods, including Transformer+OcCo~\citep{wang2021unsupervised} and Point-BERT~\citep{yu2022point}.
Our method outperforms other methods on the mean metric, as evidenced by the results. Specifically, our model achieves a mIoU that is 0.9 higher than that of the Point-BERT~\citep{yu2022point}. These results highlight the strong generalization ability of our method, particularly in scenarios with limited data.

We further evaluate the effectiveness of our method on complex and interconnected data, specifically by performing the task on the ShapeNetPart dataset. As shown in Figure~\ref{fig:seg_vis}, our predicted objects are visually similar to the ground truth, indicating that our method excels in capturing object boundaries and details.

\subsubsection{\textbf{Few-shot Classification.}}
The goal of few-shot learning is to address novel tasks with a limited number of labeled training examples by leveraging prior knowledge.
In this study, we compare the performance of our method with that of others under the conditions of $k$ classes and $m$ samples, where we sample $m$ examples for each of the $k$ classes on ModelNet40. Specifically, we present the results for the settings of $k \in \left \{5, 10\right \}$ and $m \in \left \{10, 20\right \}$ in Table~\ref{table:fewshot}.
The results demonstrate that our method consistently achieves the highest average accuracy across all four different settings, outperforming other methods by a significant margin. Specifically, our MMPT method achieves a remarkable improvement of 0.4$\%$, 0.1$\%$, 0.1$\%$, and 0.7$\%$ compared to the Point-MAE model~\citep{pang2022masked}, underscoring the robust generalization capabilities of our approach.

\subsubsection{\textbf{Indoor 3D Semantic Segmentation}} 
Furthermore, we evaluate the performance of our proposed MMPT in 3D semantic segmentation of large-scale scenes.
This task presents a significant challenge as it requires an understanding of both global semantics and local geometric details. 
The detailed quantitative results of our experiment are presented in Table~\ref{tab:indoorseg}.
Significantly, our MMPT shows a notable improvement compared to the Transformer trained from scratch, with a performance increase of 3.2$\%$ in mean accuracy (mAcc) and 4.2$\%$ in mean intersection over union (mIoU). 
This result provides evidence that our MMPT effectively enhances the capabilities of the Transformer in addressing demanding downstream tasks.
Furthermore, our MMPT surpasses other self-supervised methods, achieving the highest performance by improving the mAcc and mIoU by 1.4$\%$ and 0.3$\%$ respectively, compared to the second-best result obtained by Point-MAE.
When compared to approaches that rely on scene geometric features and colors (as illustrated in the top four methods in Table~\ref{tab:indoorseg}), our MMPT demonstrates superior performance.

\subsubsection{\textbf{Indoor 3D Object Detection}}
Furthermore, we proceed to evaluate the performance of our MMPT on the 3D object detection task, which necessitates methods with a robust understanding of large-scale scenes. To accomplish this, we conducted an experiment on the widely used real-world dataset, ScanNet V2. The results, presented in Table~\ref{tab:indoordet}, are measured in terms of $\textit{AP}_{25}$ and $\textit{AP}_{50}$. Comparing the performance of both the methods trained from scratch and the pre-training methods, our approach achieves the highest $\textit{AP}_{25}$ and $\textit{AP}_{50}$ scores. Notably, our model outperforms the second-best method by attaining a 0.2$\%$ gain in $\textit{AP}_{25}$ and a 0.2$\%$ gain in $\textit{AP}_{50}$.

\subsubsection{\textbf{3D Shape Completion.}}

\textbf{Results on PCN Dataset.} 
To evaluate the generation abilities, we fine-tune the pre-trained model on PCN dataset.
Table~\ref{tab:PCN} and the supplementary materials demonstrate that our MMPT achieves remarkable performance in terms of average Chamfer Distances~\citep{fan2017point} across all eight categories. Our method exhibits a relative improvement in averaged CD-$\ell_1$ of 65.10$\%$ and 13.06$\%$ when compared to PoinTr~\citep{yu2021pointr} and Snowflake~\citep{xiang2021snowflakenet}, resulting in a final value of 7.396. Notably, in the chair category, our MMPT achieves a remarkable CD-$\ell_1$ of 7.864, surpassing PoinTr and Snowflake by nearly 76.58$\%$ and 15.76$\%$, respectively.

To evaluate the performance of reconstructing complete shapes, we present visual comparisons of point clouds predicted by various methods on the PCN dataset in Figure~\ref{tab:PCN}. These comparisons showcase that our MMPT delivers superior visual performance compared to previous methods in missing point cloud completion tasks. Specifically, Figure~\ref{tab:PCN} demonstrates that our MMPT outperforms other methods in inferring higher-quality complete shapes in the chair category, particularly in regions of the chairs' sides and angles.

\textbf{Results on MVP Dataset.}
Moreover, we also conduct point cloud completion on MVP dataset.
Table~\ref{tab:MVP} shows that our MMPT model achieves the best outcomes in all 16 categories based on the average CD-$\ell_1$. Specifically, our MMPT model achieves an average Chamfer Distance (CD) of 6.769, which significantly outperforms PoinTr and Snowflake with average CD-$\ell_1$ of 8.070 and 7.597, respectively. In the lamp category, our method produces a significant decrease in CD-$\ell_1$, surpassing both PoinTr and Snowflake by 55.29$\%$ and 50.27$\%$, respectively.

Figure~\ref{fig:MVP} visually presents the shape results in six distinct categories, highlighting the remarkable performance of our MMPT in effectively reconstructing missing parts and capturing finer details, even with sparse input points. In the final row of Figure~\ref{fig:MVP}, other methods not only fail to reconstruct the complete structure of the motorbike but also lose its original information entirely. By contrast, our MMPT captures more intricate details and produces results with higher fidelity.

\textbf{Results on ShapeNet55 Dataset.}
Moreover, to further evaluate the generation capabilities of MMPT, we perform experiments on more challenging dataset.
Table~\ref{tab:ShapeNet55} demonstrates that MMPT achieves competitive results in terms of both F-score and CD-$\ell_2$ metrics. Notably, MMPT surpasses all other methods in terms of $\ell_1$ Chamfer Distances ($\times 10^{3}$) on average. When considering the simple, moderate, and hard settings on ShapeNet55, our MMPT model achieves $\ell_1$ Chamfer distances of 10.416, 12.455, and 17.093, respectively. These results indicate relative improvements of approximately 19.92$\%$, 13.87$\%$, and 10.05$\%$ compared to the leading baseline method, PoinTr.

The qualitative visualization results depicted in Fig.~\ref{fig:ShapeNet55/21}(a) exhibit the capacity of our MMPT model to enhance shape quality significantly across all categories of the ShapeNet55 dataset. Based on these results, we can draw the following conclusions: Our MMPT model can achieve comparable prediction accuracy, even when processing point clouds for object surfaces that are more uniform and densely distributed. Other approaches are incapable of generating shapes with more distinct structures or restoring shape details with reduced noise.

\textbf{Results on ShapeNetUnseen21 Dataset.} 
In point cloud completion, evaluating the performance on unseen objects is also necessary. Therefore, to assess the performance of our method, we conducted experiments on ShapeNetUnseen21, which is derived from ShapeNet55.
Table~\ref{tab:ShapeNet34-21} summarizes the comparison results between our MMPT model and the other nine competitive methods on the ShapeNet34 and ShapeNetUnseen21 datasets. The table indicates that our MMPT model achieves comparable or superior performance compared to PoinTr or Snowflake across all categories. As shown in Table~\ref{tab:ShapeNet34-21}, our method outperforms the second-best method, PoinTr, with relative improvements of 14.53$\%$, 1.69$\%$, 19.15$\%$, and 13.61$\%$ on the average CD of 55 categories under simple, moderate, and hard settings.
As the difficulty level of the setting increases, it is evident that the performance of all methods decreases significantly.

Figure~\ref{fig:ShapeNet55/21}(b) shows visual comparisons between our MMPT and the nine methods using the simple setting on the ShapeNetUnseen21 dataset. These comparisons reveal a significant performance gap between our approach and the baselines. Notably, our MMPT outperforms other methods, particularly when handling the incomplete point cloud representation of the basket. Our approach exhibits superior capability in recovering more precise details in the bottom-right corner of the object, while other methods fall short in achieving comparable performance, lacking the ability to capture finer details.

\subsection{Visualization of feature distributions.} 
To gain a more comprehensive understanding of the effectiveness of our method, we employ t-SNE~\citep{van2008visualizing} to visualize the learned features. Fig.~\ref{fig:tsne}(\textbf{Left}) displays our t-SNE visualization of the features learned from ModelNet40, while Fig.~\ref{fig:tsne}(\textbf{Right}) illustrates the features learned on ScanObjectNN. The visualization demonstrates that the features form numerous well-separated clusters, which confirms the effectiveness of our method.

\begin{table}[t]
\centering
\caption{Ablation study on multi-tasks of pre-training.}
\begin{tabular}{l|ccc|cc}
\toprule[1.5pt]
Model & TLR & PLR & MCL & Acc. on MN40 & Acc. on SONN \\ \midrule[1pt]
A     & \CheckmarkBold   &     &     &              93.1      &  88.0                     \\
\rowcolor[HTML]{EFEFEF}B     & \CheckmarkBold   & \CheckmarkBold   &     &  93.5                   &      88.6                \\
C     & \CheckmarkBold   &     & \CheckmarkBold   &  93.4                  &   88.3                   \\ \midrule[1pt]
\rowcolor[HTML]{EFEFEF}MMPT  & \CheckmarkBold   & \CheckmarkBold   & \CheckmarkBold   & \textbf{93.9}                   &     \textbf{91.0}                 \\ \bottomrule[1.5pt]
\end{tabular}
\label{ablation1}
\end{table}

\begin{table}[t]
\tabcolsep=0.13cm
\centering
\caption{Ablation study on the number of views.}
\begin{tabular}{l|cccccc}
\toprule[1.5pt]
Number of Views    & 1 & 2 & 3 & 4 & 5 & 6 \\ \midrule[1pt]
\rowcolor[HTML]{EFEFEF}Acc. on ModelNet40 & \textbf{93.9}   & 93.6   & 93.3  & 93.4  &  92.9  & 92.6   \\ \bottomrule[1.5pt]
\end{tabular}
\label{ablation2}
\end{table}

\begin{table}[t]
\centering
\tabcolsep=0.43cm
\caption{Ablation study on the weights of multi-tasks.}
\begin{tabular}{l|c|c}
\toprule[1.5pt]
Model & Ratio    & Acc. on ModelNet40 \\ \midrule[1pt]
\rowcolor[HTML]{EFEFEF}D     & 1:1:1    &  92.0                  \\
E     & 1:1:0.5  &  92.6                  \\
\rowcolor[HTML]{EFEFEF}F     & 1:1:0.2  &  93.4                  \\
G     & 1:1:0.01 &  93.2                \\ \midrule[1pt]
\rowcolor[HTML]{EFEFEF}MMPT  & 1:1:0.1  &  \textbf{93.9}                   \\ \bottomrule[1.5pt]
\end{tabular}
\label{ablation3}
\end{table}

\section{Ablation Study and Analysis}

\noindent\textbf{Influence on the combinations of multi-tasks.}
To gain insight into the effectiveness of multi-tasks, we conduct ablation studies on different combinations of multi-tasks.
As shown in Table~\ref{ablation1}, Model A is pre-trained only with TLR tasks, while Model B and are pre-trained under two pre-text tasks.
Our MMPT, pre-trained under the multi-model and multi-task, outperforms other models by a great margin, proving the effectiveness of our multi-task and multi-model pre-training framework.
These pre-text tasks can work corporately to enrich the representative learning of Transformer, and further improve the performance of the backbone on downstream tasks.

\noindent\textbf{Influence on the number of views.}
This study aimed to examine the impact of the image branch on the outcomes by manipulating the number of rendered 2D images. Specifically, we sought to determine how varying the number of rendered 2D images affected the results of the study.
The 2D images were rendered from various random directions. Whenever multiple rendered 2D images were utilized, we calculated the mean of all the projected features to conduct cross-modal instance discrimination.
The classification results on the ModelNet40 dataset are presented in Table~\ref{ablation2}. 
MMPT, which utilized even a single rendered 2D image, captured cross-modal correspondence and yielded superior classification results. 
Interestingly, the accuracy dropped when more than two rendered images were used, suggesting that the information gathered from the 2D image modality might have been redundant.

\noindent\textbf{Influence on the weights of multi-tasks.}
Furthermore, we conducted ablation experiments on weight combinations of different pre-training tasks.
We fixed the ratio of TLR and PLR to 1:1, as they reconstructed the point cloud from different perspectives. Additionally, we adjusted the ratio of MCL to 1, 0.5, 0.2, 0.1, and 0.01, respectively. 
As shown in Table~\ref{ablation3}, MMPT achieved better performance at a ratio of 1:1:0.1. 
This is mainly due to the trade-off between different pre-training tasks, which enables them to work collaboratively and obtain a stronger pre-trained model.

\section{Conclusion}
In summary, this paper proposes a multi-modal and multi-task pre-training framework that introduces multi-task learning to the point cloud pre-training field for the first time. 
To address the bottleneck of a single pre-training task in diverse downstream tasks, we designed three pre-training tasks: TLR, PLR, and MCL. 
These three pre-training tasks work collaboratively to obtain a pre-trained model with rich representation capabilities. 
The pre-trained model achieved satisfactory performance on five downstream tasks. 
In the future, more multi-task pre-training models for specific downstream tasks will be developed based on our work to promote the development of pre-training in the 3D field with low annotation and high transfer performance.

\bibliographystyle{unsrtnat}
\bibliography{ref}  






\appendix

The supplementary materials provide an extensive analysis of the quantitative results achieved by our MMPT approach in comparison to other state-of-the-art methods. The evaluation includes performance metrics across multiple categories and datasets, such as PCN, MVP, ShapeNet55, ShapeNet34, and ShapeNetUnseen21, as depicted in Table~\ref{PCN-1}-Table~\ref{21}.

\section{Datasets}

\textbf{ModelNet40}~\citep{wu20153d} dataset comprises 12,311 CAD models from 40 object categories, with 9,843 samples used for training and 2,468 samples for testing. We follow previous works and use 1024 points with coordinate information as the input~\citep{yu2022point, lu20223dctn, gao2022lft}.

\textbf{The ScanObjectNN}~\citep{uy2019revisiting} dataset is a challenging dataset that consists of 15,000 objects from 15 categories based on scanned indoor scene data. It is divided into 11,416 instances for the training set and 2,882 for validation. We evaluate our experiments on three variants: OBJ-BG, OBJ-ONLY, and PB-T50-RS, which are consistent with previous works~\citep{huang2022dual, liu2023self}.

\textbf{The ShapeNetPart}~\citep{yi2016scalable} dataset consists of 16 different categories and 16,881 3D objects. The training set comprises 14,007 samples, while the remaining 2,874 samples are used for validation. The mIoU metric is employed to evaluate the performance of different methods, providing a comprehensive and detailed understanding of their effectiveness.

\textbf{PCN dataset}\citep{yuan2018pcn} is derived from the ShapeNet dataset and comprises 8 types of objects. Each complete shape is represented by 16384 points, which are uniformly sampled from the surface of the original CAD model.

\textbf{MVP dataset}\citep{pan2021variational} expands the existing 8 categories in the PCN dataset by introducing an additional 8 categories, including bed, bench, bookshelf, bus, guitar, motorbike, pistol, and skateboard, resulting in a comprehensive set of high-quality partial and complete point clouds.

\textbf{ShapeNet55/34/Unseen21.} ShapeNet55\citep{yu2021pointr} dataset comprises 57,448 synthetic 3D shapes of 55 categories and is divided into ShapeNet34\citep{yu2021pointr} and ShapeNetUnseen21 to better assess the model's generalization ability.

\textbf{S3DIS dataset} also known as the Stanford Large-Scale 3D Indoor Spaces dataset~\citep{armeni20163d}, offers instance-level semantic segmentation for six expansive indoor areas. These areas consist of a total of 271 rooms and encompass 13 distinct semantic categories. Consistent with established conventions, we designated area 5 specifically for testing purposes, while utilizing the remaining areas for training our models.

\textbf{Indoor Detection.}
The benchmark widely recognized for 3D object detection is ScanNet V2~\citep{dai2017scannet}, which comprises 1,513 indoor scenes and encompasses 18 distinct object classes. To ensure consistency, we adopt the evaluation procedure established by VoteNet~\citep{qi2019deep}, which calculates the mean average precision for two threshold values: 0.25 (mAP@0.25) and 0.5 (mAP@0.5). These metrics allow us to effectively evaluate the performance of our MMPT.

\section{Fine-tuning Setups}
We conduct experiments on two benchmarks to evaluate our object classification method, ModelNet40~\citep{wu20153d} and ScanObjectNN~\citep{uy2019revisiting}, where we perform synthetic object classification on ModelNet40. ModelNet40 consists of 12,331 meshed models from 40 object categories, with 9,843 training meshes and 2,468 testing meshes, from which the points are sampled. ScanObjectNN is a more challenging 3D point cloud classification benchmark dataset, containing 2,880 occluded objects from 15 categories captured from real indoor scenes. We follow the same settings as~\citep{qi2017pointnet,qi2017pointnet++} for fine-tuning. For PointNet, we utilize the Adam optimizer with an initial learning rate of 1e-3, and the learning rate is decayed by 0.7 every 20 epochs with the minimum value of 1e-5. For DGCNN, we use the SGD optimizer with a momentum of 0.9 and a weight decay of 1e-4. The learning rate starts from 0.1 and then decays using cosine annealing with the minimum value of 1e-3. We also apply dropout in the fully connected layers before the softmax output layer, with the dropout rate set to 0.7 for PointNet and 0.5 for DGCNN. We train all the models for 200 epochs with a batch size of 32. 

For the fine-grained 3D recognition task of part segmentation, we use ShapeNetPart~\citep{yi2016scalable}, which comprises 16,881 objects of 2,048 points from 16 categories with 50 parts in total. Similar to PointNet~\citep{qi2017pointnet}, we sample 2,048 points from each model. For PointNet, we adopt the Adam optimizer with an initial learning rate of 1e-3, and the learning rate is decayed by 0.5 every 20 epochs with the minimum value of 1e-5. For DGCNN, we use the SGD optimizer with a momentum of 0.9 and a weight decay of 1e-4. The learning rate starts from 0.1 and then decays using cosine annealing with the minimum value of 1e-3. We train the models for 250 epochs with a batch size of 16.

For point cloud completion task, we utilize a standard Transformer encoder and a powerful Transformer-based decoder devised in SnowflakeNet \citep{xiang2021snowflakenet}. We fine-tune our model on the point cloud completion benchmarks for 200 epochs.




\newpage
\begin{table*}[ht]
\centering
\caption{Point cloud completion on PCN in terms of F-score$@1\%$ (higher is better).}
\label{PCN-1}
\resizebox{\textwidth}{!}{%
\begin{tabular}{cccccccccc}
\hline
\rowcolor[HTML]{EFEFEF}F-score$@1\%$  & Airplane & Cabinet & Car & Chair & Lamp & Sofa & Table & watercraft & Avg. \\ \hline
ASFM~\citep{xia2021asfm} & 0.738 & 0.330 & 0.409 & 0.399 & 0.496 & 0.331 & 0.411 & 0.556 & 0.459 \\
\rowcolor[HTML]{EFEFEF} CRN~\citep{wang2021cascaded} & 0.804 & 0.439 & 0.486 & 0.505 & 0.533 & 0.409 & 0.606 & 0.607 & 0.549 \\
ECG~\citep{pan2020ecg} & 0.870 & 0.605 & 0.675 & 0.635 & 0.698 & 0.502 & 0.730 & 0.759 & 0.684 \\
\rowcolor[HTML]{EFEFEF} FoldingNet~\citep{yang2018foldingnet} & 0.723 & 0.310 & 0.488 & 0.307 & 0.332 & 0.280 & 0.457 & 0.449 & 0.418 \\
GRNet~\citep{xie2020grnet} & 0.711 & 0.447 & 0.507 & 0.483 & 0.609 & 0.415 & 0.556 & 0.603 & 0.541 \\
\rowcolor[HTML]{EFEFEF} PCN~\citep{yuan2018pcn} & 0.831 & 0.508 & 0.649 & 0.506 & 0.522 & 0.433 & 0.621 & 0.644 & 0.589 \\
TopNet~\citep{tchapmi2019topnet} & 0.760 & 0.321 & 0.496 & 0.342 & 0.345 & 0.313 & 0.504 & 0.461 & 0.443 \\ 
\rowcolor[HTML]{EFEFEF} PoinTr~\citep{yu2021pointr} & 0.704 & 0.435 & 0.549 & 0.447 & 0.528 & 0.384 & 0.602 & 0.570 & 0.527 \\
SnowflakeNet~\citep{xiang2021snowflakenet} & 0.897 & 0.648 & 0.697 & 0.705 & 0.790 & 0.604 & 0.820 & 0.787 & 0.743 \\\hline
\rowcolor[HTML]{EFEFEF} MMPT & 0.929 & 0.649 & 0.712 & 0.751 & 0.856 & 0.635 & 0.835 & 0.836 & 0.775 \\ \hline
\end{tabular}%
}
\end{table*}

\begin{table*}[ht]
\centering
\caption{Point cloud completion on PCN in terms of L2 Chamfer distance $\times 10^{3}$ (lower is better).}
\resizebox{\textwidth}{!}{%
\begin{tabular}{cccccccccc}
\hline
\rowcolor[HTML]{EFEFEF} CD-$\ell2$($\times10^{3}$) & Airplane & Cabinet & Car & Chair & Lamp & Sofa & Table & watercraft & Avg. \\ \hline
ASFM~\citep{xia2021asfm} & 0.334 & 1.158 & 0.597 & 1.030 & 1.187 & 1.180 & 1.212 & 0.647 & 0.918 \\
\rowcolor[HTML]{EFEFEF} CRN~\citep{wang2021cascaded} & 0.351 & 0.704 & 0.526 & 0.653 & 0.842 & 0.896 & 0.570 & 0.480 & 0.628 \\
ECG~\citep{pan2020ecg} & 0.169 & 0.502 & 0.253 & 0.459 & 0.570 & 0.632 & 0.380 & 0.299 & 0.408 \\
\rowcolor[HTML]{EFEFEF} FoldingNet~\citep{yang2018foldingnet} & 0.264 & 0.665 & 0.355 & 0.719 & 0.733 & 0.687 & 0.624 & 0.514 & 0.570 \\
GRNet~\citep{xie2020grnet} & 0.405 & 0.786 & 0.462 & 0.808 & 0.689 & 0.950 & 0.672 & 0.525 & 0.662 \\
\rowcolor[HTML]{EFEFEF} PCN~\citep{yuan2018pcn} & 0.213 & 0.633 & 0.313 & 0.638 & 0.750 & 0.784 & 0.573 & 0.433 & 0.542 \\
TopNet~\citep{tchapmi2019topnet} & 0.240 & 0.712 & 0.404 & 0.856 & 0.710 & 0.771 & 0.595 & 0.504 & 0.599 \\ 
\rowcolor[HTML]{EFEFEF} PoinTr~\citep{yu2021pointr} & 0.342 & 0.682 & 0.415 & 0.704 & 0.616 & 0.815 & 0.526 & 0.415 & 0.564 \\
SnowflakeNet~\citep{xiang2021snowflakenet} & 0.131 & 0.439 & 0.246 & 0.343 & 0.343 & 0.493 & 0.294 & 0.196 & 0.311 \\\hline
\rowcolor[HTML]{EFEFEF} MMPT & 0.094 & 0.441 & 0.242 & 0.276 & 0.195 & 0.438 & 0.228 & 0.168 & 0.260 \\ \hline
\end{tabular}%
}
\end{table*}

\begin{table*}[htbp]
\centering
\tabcolsep=0.09cm
\caption{Point cloud completion performance on MVP dataset in terms of F-Score$@1\%$ (higher is better).}
\resizebox{\textwidth}{!}{%
\begin{tabular}{cccccccccccccccccc}
\hline
\rowcolor[HTML]{EFEFEF} F-score$@1\%$ & Chair & Table & Sofa & Cabinet & Lamp & Car & Airplane & Watercraft & Bed & Bench & Bookshelf & Bus & Guitar & Motorbike & Pistol & Skateboard & Avg. \\ \hline
ASFM~\citep{xia2021asfm} & 0.517 & 0.587 & 0.473 & 0.486 & 0.593 & 0.533 & 0.857 & 0.632 & 0.438 & 0.674 & 0.461 & 0.673 & 0.886 & 0.626 & 0.749 & 0.822 & 0.605 \\
\rowcolor[HTML]{EFEFEF} CRN~\citep{wang2021cascaded} & 0.656 & 0.726 & 0.639 & 0.674 & 0.665 & 0.734 & 0.891 & 0.732 & 0.527 & 0.704 & 0.626 & 0.718 & 0.644 & 0.623 & 0.594 & 0.700 & 0.696 \\
ECG~\citep{pan2020ecg} & 0.682 & 0.759 & 0.649 & 0.681 & 0.714 & 0.726 & 0.843 & 0.750 & 0.644 & 0.804 & 0.707 & 0.814 & 0.779 & 0.784 & 0.841 & 0.900 & 0.740 \\
\rowcolor[HTML]{EFEFEF} FoldingNet~\citep{yang2018foldingnet} & 0.334 & 0.518 & 0.392 & 0.496 & 0.320 & 0.572 & 0.739 & 0.515 & 0.309 & 0.604 & 0.407 & 0.705 & 0.848 & 0.511 & 0.706 & 0.782 & 0.516 \\
GRNet~\citep{xie2020grnet} & 0.501 & 0.579 & 0.497 & 0.550 & 0.536 & 0.613 & 0.825 & 0.635 & 0.467 & 0.645 & 0.504 & 0.679 & 0.815 & 0.668 & 0.740 & 0.778 & 0.609 \\
\rowcolor[HTML]{EFEFEF} PCN~\citep{yuan2018pcn} & 0.387 & 0.538 & 0.453 & 0.570 & 0.357 & 0.641 & 0.785 & 0.529 & 0.334 & 0.559 & 0.461 & 0.746 & 0.879 & 0.607 & 0.705 & 0.816 & 0.559 \\
TopNet~\citep{tchapmi2019topnet} & 0.342 & 0.501 & 0.364 & 0.420 & 0.336 & 0.511 & 0.747 & 0.485 & 0.327 & 0.595 & 0.347 & 0.618 & 0.812 & 0.532 & 0.638 & 0.754 & 0.492 \\ 
\rowcolor[HTML]{EFEFEF} PoinTr~\citep{yu2021pointr} & 0.725 & 0.795 & 0.717 & 0.729 & 0.737 & 0.751 & 0.916 & 0.791 & 0.654 & 0.853 & 0.728 & 0.842 & 0.952 & 0.796 & 0.868 & 0.924 & 0.784 \\
SnowflakeNet~\citep{xiang2021snowflakenet} & 0.762 & 0.840 & 0.742 & 0.763 & 0.782 & 0.766 & 0.934 & 0.797 & 0.709 & 0.874 & 0.787 & 0.848 & 0.969 & 0.833 & 0.893 & 0.943 & 0.813 \\\hline
\rowcolor[HTML]{EFEFEF} MMPT & 0.760 & 0.816 & 0.713 & 0.685 & 0.838 & 0.715 & 0.939 & 0.814 & 0.714 & 0.874 & 0.775 & 0.814 & 0.967 & 0.815 & 0.883 & 0.936 & 0.801 \\ \hline
\end{tabular}%
}
\end{table*}

\begin{table*}[htbp]
\centering
\tabcolsep=0.09cm
\caption{Point cloud completion performance on MVP dataset in terms of L2 Chamfer distance $\times 10^{3}$ (lower is better).}
\resizebox{\textwidth}{!}{%
\begin{tabular}{cccccccccccccccccc}
\hline
\rowcolor[HTML]{EFEFEF} CD-$\ell2$($\times10^{3}$) & Chair & Table & Sofa & Cabinet & Lamp & Car & Airplane & Watercraft & Bed & Bench & Bookshelf & Bus & Guitar & Motorbike & Pistol & Skateboard & Avg. \\ \hline
ASFM~\citep{xia2021asfm} & 0.879 & 0.972 & 0.685 & 0.597 & 1.336 & 0.405 & 0.234 & 0.669 & 1.567 & 0.617 & 1.087 & 0.345 & 0.109 & 0.328 & 0.354 & 0.357 & 0.691 \\
\rowcolor[HTML]{EFEFEF} CRN~\citep{wang2021cascaded} & 0.625 & 0.724 & 0.488 & 0.404 & 0.954 & 0.264 & 0.190 & 0.436 & 2.139 & 0.853 & 1.018 & 0.405 & 0.899 & 0.525 & 1.974 & 0.758 & 0.651 \\
ECG~\citep{pan2020ecg} & 0.522 & 0.506 & 0.429 & 0.385 & 0.881 & 0.243 & 0.185 & 0.394 & 0.859 & 0.375 & 0.520 & 0.216 & 0.179 & 0.209 & 0.235 & 0.156 & 0.418 \\
\rowcolor[HTML]{EFEFEF} FoldingNet~\citep{yang2018foldingnet} & 0.754 & 0.572 & 0.570 & 0.412 & 1.536 & 0.314 & 0.339 & 0.563 & 1.022 & 0.636 & 0.561 & 0.255 & 0.120 & 0.358 & 0.268 & 1.267 & 0.615 \\
GRNet~\citep{xie2020grnet} & 0.915 & 0.902 & 0.670 & 0.558 & 1.431 & 0.366 & 0.336 & 0.530 & 1.302 & 0.610 & 0.890 & 0.374 & 0.163 & 0.302 & 0.276 & 0.701 & 0.679 \\
\rowcolor[HTML]{EFEFEF} PCN~\citep{yuan2018pcn} & 1.240 & 1.119 & 0.795 & 0.543 & 2.363 & 0.373 & 0.374 & 0.856 & 1.938 & 0.987 & 1.178 & 0.323 & 0.120 & 0.431 & 0.558 & 0.347 & 0.902 \\
TopNet~\citep{tchapmi2019topnet} & 0.741 & 0.674 & 0.599 & 0.508 & 1.189 & 0.370 & 0.286 & 0.637 & 1.023 & 0.493 & 0.779 & 0.316 & 0.153 & 0.372 & 0.321 & 0.218 & 0.584 \\ 
\rowcolor[HTML]{EFEFEF} PoinTr~\citep{yu2021pointr} & 0.398 & 0.444 & 0.381 & 0.374 & 0.615 & 0.240 & 0.134 & 0.267 & 0.818 & 0.272 & 0.439 & 0.193 & 0.056 & 0.188 & 0.185 & 0.107 & 0.338 \\
SnowflakeNet~\citep{xiang2021snowflakenet} & 0.405 & 0.391 & 0.388 & 0.403 & 0.690 & 0.238 & 0.113 & 0.285 & 0.736 & 0.264 & 0.380 & 0.206 & 0.042 & 0.167 & 0.170 & 0.121 & 0.338 \\\hline
\rowcolor[HTML]{EFEFEF} MMPT & 0.289 & 0.269 & 0.284 & 0.328 & 0.245 & 0.235 & 0.072 & 0.200 & 0.442 & 0.176 & 0.309 & 0.184 & 0.044 & 0.150 & 0.130 & 0.103 & 0.228 \\ \hline
\end{tabular}%
}
\end{table*}
\onecolumn

\newpage
\setlength\tabcolsep{4pt}{
\tiny
\begin{longtable}[c]{ccccccccccccc|ccc}
\caption{Quantitative results of ASFM, CRN, ECG, FoldingNet, and MMPT on ShapeNet55.}
\\
\hline
 & \multicolumn{3}{c}{ASFM} & \multicolumn{3}{c}{CRN} & \multicolumn{3}{c}{ECG} & \multicolumn{3}{c|}{FoldingNet} & \multicolumn{3}{c}{MMPT} \\ \cline{2-16} 
\multirow{-2}{*}{\begin{tabular}[c]{@{}c@{}}F-Score/CD-$\ell1$($\times$\\$10^{3}$)/CD-$\ell2$($\times10^{3}$)\end{tabular}} & Simple & Moderate & Hard & Simple & Moderate & Hard & Simple & Moderate & Hard & Simple & Moderate & Hard & Simple & Moderate & Hard \\ \hline
\endfirsthead
\multicolumn{16}{c}%
{{\bfseries Table \thetable\ continued from previous page}} 
\vspace{0.1cm}
\\
\hline
 & \multicolumn{3}{c}{ASFM~\citep{xia2021asfm}} & \multicolumn{3}{c}{CRN~\citep{wang2021cascaded}} & \multicolumn{3}{c}{ECG~\citep{pan2020ecg}} & \multicolumn{3}{c|}{FoldingNet~\citep{yang2018foldingnet}} & \multicolumn{3}{c}{MMPT} \\ \cline{2-16} 
\multirow{-2}{*}{\begin{tabular}[c]{@{}c@{}}F-Score/CD-$\ell1$($\times$\\$10^{3}$)/CD-$\ell2$($\times10^{3}$)\end{tabular}} & Simple & Moderate & Hard & Simple & Moderate & Hard & Simple & Moderate & Hard & Simple & Moderate & Hard & Simple & Moderate & Hard \\ \hline
\endhead
\hline
\endfoot
\endlastfoot
\rowcolor[HTML]{EFEFEF} 
\cellcolor[HTML]{EFEFEF} & 0.355 & 0.345 & 0.302 & 0.370 & 0.351 & 0.302 & 0.490 & 0.441 & 0.352 & 0.108 & 0.088 & 0.056 & 0.675 & 0.593 & 0.427 \\
\rowcolor[HTML]{EFEFEF} 
\cellcolor[HTML]{EFEFEF} & 18.681 & 21.224 & 25.969 & 19.947 & 23.496 & 30.030 & 18.152 & 22.088 & 27.676 & 27.366 & 30.365 & 35.458 & 8.974 & 12.529 & 20.170 \\
\rowcolor[HTML]{EFEFEF} 
\multirow{-3}{*}{\cellcolor[HTML]{EFEFEF}Microphone} & 1.909 & 2.683 & 3.957 & 2.109 & 3.291 & 5.450 & 2.368 & 3.249 & 5.122 & 3.283 & 4.193 & 5.922 & 1.017 & 1.976 & 4.323 \\
 & 0.195 & 0.187 & 0.170 & 0.142 & 0.141 & 0.135 & 0.257 & 0.258 & 0.233 & 0.063 & 0.056 & 0.043 & 0.382 & 0.367 & 0.312 \\
 & 20.186 & 21.456 & 25.502 & 23.414 & 24.675 & 28.686 & 17.797 & 19.838 & 24.755 & 26.643 & 28.007 & 31.703 & 11.188 & 12.883 & 17.438 \\
\multirow{-3}{*}{Stove} & 1.319 & 1.560 & 2.378 & 1.656 & 1.960 & 2.935 & 1.088 & 1.459 & 2.373 & 2.069 & 2.369 & 3.196 & 0.596 & 0.913 & 1.880 \\
\rowcolor[HTML]{EFEFEF} 
\cellcolor[HTML]{EFEFEF} & 0.185 & 0.170 & 0.166 & 0.181 & 0.179 & 0.168 & 0.288 & 0.292 & 0.248 & 0.041 & 0.036 & 0.028 & 0.412 & 0.361 & 0.282 \\
\rowcolor[HTML]{EFEFEF} 
\cellcolor[HTML]{EFEFEF} & 22.475 & 24.088 & 31.260 & 24.343 & 26.074 & 34.238 & 21.372 & 24.061 & 35.406 & 34.673 & 36.760 & 43.769 & 12.877 & 18.554 & 26.160 \\
\hline
\end{longtable}

\newpage
\setlength\tabcolsep{2.5pt}{
\tiny
\begin{longtable}[c]{cccccccccccccccc|ccc}
\caption{Quantitative results of GRNet, PCN, TopNet, PoinTr, Snowflake, and MMPT on ShapeNet55.}
\\
\hline
 & \multicolumn{3}{c}{GRNet} & \multicolumn{3}{c}{PCN} & \multicolumn{3}{c}{TopNet} & \multicolumn{3}{c}{PoinTr} & \multicolumn{3}{c|}{Snowflake} & \multicolumn{3}{c}{MMPT} \\ \cline{2-19} 
\multirow{-2}{*}{\begin{tabular}[c]{@{}c@{}}F-Score/CD-$\ell1$($\times$\\$10^{3}$)/CD-$\ell2$($\times10^{3}$)\end{tabular}} & Simple & Moderate & Hard & Simple & Moderate & Hard & Simple & Moderate & Hard & Simple & Moderate & Hard & Simple & Moderate & Hard & Simple & Moderate & Hard \\ \hline
\endfirsthead
\multicolumn{19}{c}%
{{\bfseries Table \thetable\ continued from previous page}} 
\vspace{0.1cm}
\\
\hline
 & \multicolumn{3}{c}{GRNet~\citep{xie2020grnet}} & \multicolumn{3}{c}{PCN~\citep{yuan2018pcn}} & \multicolumn{3}{c}{TopNet~\citep{tchapmi2019topnet}} & \multicolumn{3}{c}{PoinTr~\citep{yu2021pointr}} & \multicolumn{3}{c|}{SnowflakeNet~\citep{xiang2021snowflakenet}} & \multicolumn{3}{c}{MMPT} \\ \cline{2-19} 
\multirow{-2}{*}{\begin{tabular}[c]{@{}c@{}}F-Score/CD-$\ell1$($\times$\\$10^{3}$)/CD-$\ell2$($\times10^{3}$)\end{tabular}} & Simple & Moderate & Hard & Simple & Moderate & Hard & Simple & Moderate & Hard & Simple & Moderate & Hard & Simple & Moderate & Hard & Simple & Moderate & Hard \\ \hline
\endhead
\hline
\endfoot
\endlastfoot
\rowcolor[HTML]{EFEFEF} 
\cellcolor[HTML]{EFEFEF} & 0.468 & 0.437 & 0.356 & 0.260 & 0.240 & 0.198 & 0.156 & 0.147 & 0.081 & 0.629 & 0.560 & 0.437 & 0.609 & 0.540 & 0.412 & 0.675 & 0.593 & 0.427 \\
\rowcolor[HTML]{EFEFEF} 
\cellcolor[HTML]{EFEFEF} & 15.754 & 20.357 & 28.788 & 26.149 & 29.819 & 35.711 & 31.040 & 34.470 & 57.314 & 11.199 & 15.443 & 23.434 & 11.829 & 15.222 & 21.320 & 8.974 & 12.529 & 20.170 \\
\rowcolor[HTML]{EFEFEF} 
\multirow{-3}{*}{\cellcolor[HTML]{EFEFEF}Microphone} & 1.312 & 2.808 & 5.995 & 3.615 & 4.883 & 7.026 & 4.526 & 5.478 & 20.796 & 0.976 & 2.050 & 4.514 & 0.914 & 1.685 & 3.102 & 1.017 & 1.976 & 4.323 \\
 & 0.171 & 0.169 & 0.163 & 0.109 & 0.101 & 0.091 & 0.079 & 0.071 & 0.061 & 0.407 & 0.398 & 0.349 & 0.326 & 0.296 & 0.249 & 0.382 & 0.367 & 0.312 \\
 & 21.264 & 22.775 & 26.869 & 24.659 & 25.837 & 29.642 & 28.619 & 30.414 & 34.814 & 13.644 & 15.061 & 19.821 & 14.706 & 16.508 & 20.762 & 11.188 & 12.883 & 17.438 \\
\multirow{-3}{*}{Stove} & 1.271 & 1.624 & 2.591 & 1.844 & 2.105 & 2.992 & 2.461 & 2.847 & 3.938 & 0.743 & 1.052 & 1.980 & 0.720 & 1.003 & 1.758 & 0.596 & 0.913 & 1.880 \\
\rowcolor[HTML]{EFEFEF} 
\cellcolor[HTML]{EFEFEF} & 0.247 & 0.245 & 0.229 & 0.125 & 0.122 & 0.107 & 0.076 & 0.071 & 0.058 & 0.433 & 0.414 & 0.337 & 0.357 & 0.321 & 0.265 & 0.412 & 0.361 & 0.282 \\
\rowcolor[HTML]{EFEFEF} 
\cellcolor[HTML]{EFEFEF} & 20.222 & 22.659 & 29.192 & 28.030 & 29.324 & 35.686 & 35.102 & 37.333 & 47.208 & 14.697 & 17.663 & 26.731 & 16.114 & 18.317 & 25.224 & 12.877 & 18.554 & 26.160 \\
\hline
\end{longtable}
}

\newpage
\setlength\tabcolsep{3.5pt}{
\scriptsize
\begin{longtable}[c]{ccccccccccccc|ccc}
\caption{Quantitative results of ASFM, CRN, ECG, FoldingNet, and MMPT on ShapeNet34.}
\vspace{0.35cm}
\\
\hline
 & \multicolumn{3}{c}{ASFM} & \multicolumn{3}{c}{CRN} & \multicolumn{3}{c}{ECG} & \multicolumn{3}{c|}{FoldingNet} & \multicolumn{3}{c}{MMPT} \\ \cline{2-16} 
\multirow{-2}{*}{\begin{tabular}[c]{@{}c@{}}F-Score/CD-$\ell1$($\times$\\$10^{3}$)/CD-$\ell2$($\times10^{3}$)\end{tabular}} & Simple & Moderate & Hard & Simple & Moderate & Hard & Simple & Moderate & Hard & Simple & Moderate & Hard & Simple & Moderate & Hard \\ \hline
\endfirsthead
\multicolumn{16}{c}%
{{\bfseries Table \thetable\ continued from previous page}} 
\vspace{0.1cm}
\\
\hline
 & \multicolumn{3}{c}{ASFM~\citep{xia2021asfm}} & \multicolumn{3}{c}{CRN~\citep{wang2021cascaded}} & \multicolumn{3}{c}{ECG~\citep{pan2020ecg}} & \multicolumn{3}{c|}{FoldingNet~\citep{yang2018foldingnet}} & \multicolumn{3}{c}{MMPT} \\ \cline{2-16} 
\multirow{-2}{*}{\begin{tabular}[c]{@{}c@{}}F-Score/CD-$\ell1$($\times$\\$10^{3}$)/CD-$\ell2$($\times10^{3}$)\end{tabular}} & Simple & Moderate & Hard & Simple & Moderate & Hard & Simple & Moderate & Hard & Simple & Moderate & Hard & Simple & Moderate & Hard \\ \hline
\endhead
\hline
\endfoot
\endlastfoot
\rowcolor[HTML]{EFEFEF} 
\cellcolor[HTML]{EFEFEF} & 0.132 & 0.131 & 0.117 & 0.098 & 0.100 & 0.098 & 0.370 & 0.382 & 0.347 & 0.062 & 0.058 & 0.041 & 0.267 & 0.253 & 0.215 \\
\rowcolor[HTML]{EFEFEF} 
\cellcolor[HTML]{EFEFEF} & 22.107 & 23.654 & 28.270 & 25.004 & 26.704 & 32.221 & 16.527 & 18.563 & 24.481 & 27.615 & 29.090 & 34.714 & 13.616 & 15.771 & 21.123 \\
\rowcolor[HTML]{EFEFEF} 
\multirow{-3}{*}{\cellcolor[HTML]{EFEFEF}Trash bin} & 1.388 & 1.699 & 2.715 & 1.688 & 2.172 & 3.630 & 0.915 & 1.350 & 2.673 & 2.168 & 2.509 & 3.837 & 0.817 & 1.218 & 2.500 \\
 & 0.603 & 0.557 & 0.496 & 0.607 & 0.578 & 0.514 & 0.802 & 0.731 & 0.600 & 0.363 & 0.337 & 0.306 & 0.861 & 0.779 & 0.625 \\
 & 10.983 & 11.600 & 13.222 & 11.032 & 11.731 & 13.373 & 7.917 & 8.940 & 11.191 & 14.369 & 15.489 & 16.506 & 5.095 & 6.464 & 9.018 \\
\multirow{-3}{*}{Knife} & 0.506 & 0.549 & 0.751 & 0.533 & 0.612 & 0.860 & 0.348 & 0.430 & 0.690 & 0.827 & 0.943 & 1.056 & 0.203 & 0.360 & 0.701 \\
\rowcolor[HTML]{EFEFEF} 
\cellcolor[HTML]{EFEFEF} & 0.367 & 0.359 & 0.320 & 0.242 & 0.241 & 0.229 & 0.553 & 0.551 & 0.464 & 0.226 & 0.218 & 0.186 & 0.499 & 0.504 & 0.453 \\
\rowcolor[HTML]{EFEFEF} 
\cellcolor[HTML]{EFEFEF} & 14.710 & 15.272 & 17.574 & 17.316 & 17.967 & 20.254 & 11.573 & 12.596 & 16.000 & 19.301 & 19.826 & 22.228 & 9.103 & 10.197 & 13.296 \\
\hline
\end{longtable}}

\newpage
\setlength\tabcolsep{2pt}{
\Huge
\tiny
\begin{longtable}[c]{cccccccccccccccc|ccc}
\caption{Quantitative completion results of GRNet, PCN, TopNet, PoinTr, Snowflake, and MMPT on ShapeNet34.}
\vspace{0.35cm}
\\
\hline
 & \multicolumn{3}{c}{GRNet~\citep{xie2020grnet}} & \multicolumn{3}{c}{PCN~\citep{yuan2018pcn}} & \multicolumn{3}{c}{TopNet~\citep{tchapmi2019topnet}} & \multicolumn{3}{c}{PoinTr~\citep{yu2021pointr}} & \multicolumn{3}{c|}{SnowflakeNet~\citep{xiang2021snowflakenet}} & \multicolumn{3}{c}{MMPT} \\ \cline{2-19} 
\multirow{-2}{*}{\begin{tabular}[c]{@{}c@{}}F-Score/CD-$\ell1$($\times$\\$10^{3}$)/CD-$\ell2$($\times10^{3}$)\end{tabular}} & Simple & Moderate & Hard & Simple & Moderate & Hard & Simple & Moderate & Hard & Simple & Moderate & Hard & Simple & Moderate & Hard & Simple & Moderate & Hard \\ \hline
\endfirsthead
\multicolumn{19}{c}%
{{\bfseries Table \thetable\ continued from previous page}} 
\vspace{0.1cm}
\\
\hline
 & \multicolumn{3}{c}{GRNet} & \multicolumn{3}{c}{PCN} & \multicolumn{3}{c}{TopNet} & \multicolumn{3}{c}{PoinTr} & \multicolumn{3}{c|}{Snowflake} & \multicolumn{3}{c}{MMPT} \\ \cline{2-19} 
\multirow{-2}{*}{\begin{tabular}[c]{@{}c@{}}F-Score/CD-$\ell1$($\times$\\$10^{3}$)/CD-$\ell2$($\times10^{3}$)\end{tabular}} & Simple & Moderate & Hard & Simple & Moderate & Hard & Simple & Moderate & Hard & Simple & Moderate & Hard & Simple & Moderate & Hard & Simple & Moderate & Hard \\ \hline
\endhead
\hline
\endfoot
\endlastfoot
\rowcolor[HTML]{EFEFEF} 
\cellcolor[HTML]{EFEFEF} & 0.098 & 0.099 & 0.100 & 0.075 & 0.073 & 0.062 & 0.061 & 0.056 & 0.043 & 0.342 & 0.334 & 0.305 & 0.225 & 0.199 & 0.167 & 0.267 & 0.253 & 0.215 \\
\rowcolor[HTML]{EFEFEF} 
\cellcolor[HTML]{EFEFEF} & 24.351 & 25.924 & 29.998 & 25.674 & 27.126 & 32.316 & 27.413 & 28.740 & 33.682 & 15.840 & 17.593 & 22.943 & 17.922 & 20.237 & 25.853 & 13.616 & 15.771 & 21.123 \\
\rowcolor[HTML]{EFEFEF} 
\multirow{-3}{*}{\cellcolor[HTML]{EFEFEF}Trash bin} & 1.537 & 1.939 & 2.948 & 1.787 & 2.180 & 3.478 & 2.040 & 2.349 & 3.594 & 0.897 & 1.279 & 2.416 & 0.972 & 1.371 & 2.517 & 0.817 & 1.218 & 2.500 \\
 & 0.642 & 0.598 & 0.507 & 0.538 & 0.514 & 0.495 & 0.441 & 0.420 & 0.372 & 0.797 & 0.739 & 0.622 & 0.789 & 0.726 & 0.593 & 0.861 & 0.779 & 0.625 \\
 & 10.923 & 11.983 & 14.430 & 12.694 & 13.297 & 14.123 & 13.421 & 13.992 & 15.369 & 6.827 & 8.163 & 10.821 & 7.664 & 8.680 & 11.277 & 5.095 & 6.464 & 9.018 \\
\multirow{-3}{*}{Knife} & 0.514 & 0.641 & 1.073 & 0.692 & 0.738 & 0.919 & 0.655 & 0.732 & 0.935 & 0.260 & 0.451 & 0.794 & 0.292 & 0.417 & 0.707 & 0.203 & 0.360 & 0.701 \\
\rowcolor[HTML]{EFEFEF} 
\cellcolor[HTML]{EFEFEF} & 0.249 & 0.247 & 0.238 & 0.225 & 0.221 & 0.201 & 0.209 & 0.205 & 0.186 & 0.490 & 0.496 & 0.452 & 0.432 & 0.417 & 0.375 & 0.499 & 0.504 & 0.453 \\
\rowcolor[HTML]{EFEFEF} 
\cellcolor[HTML]{EFEFEF} & 16.999 & 17.768 & 19.970 & 18.043 & 18.531 & 20.828 & 18.656 & 19.147 & 21.138 & 10.685 & 11.237 & 14.024 & 12.027 & 13.017 & 16.100 & 9.103 & 10.197 & 13.296 \\
\hline
\end{longtable}}

\newpage
\setlength\tabcolsep{2pt}{
\Huge
\tiny
\begin{longtable}[c]{ccccccccccccc|ccc}
\caption{Quantitative completion results of ASFM, CRN, ECG, FoldingNet, and MMPT on ShapeNetUnseen21.}
\vspace{0.35cm}
\\
\hline
 & \multicolumn{3}{c}{ASFM~\citep{xia2021asfm}} & \multicolumn{3}{c}{CRN~\citep{wang2021cascaded}} & \multicolumn{3}{c}{ECG~\citep{pan2020ecg}} & \multicolumn{3}{c|}{FoldingNet~\citep{yang2018foldingnet}} & \multicolumn{3}{c}{MMPT} \\ \cline{2-16} 
\multirow{-2}{*}{\begin{tabular}[c]{@{}c@{}}F-Score/CD-$\ell1$($\times$\\$10^{3}$)/CD-$\ell2$($\times10^{3}$)\end{tabular}} & Simple & Moderate & Hard & Simple & Moderate & Hard & Simple & Moderate & Hard & Simple & Moderate & Hard & Simple & Moderate & Hard \\ \hline
\endfirsthead
\multicolumn{16}{c}%
{{\bfseries Table \thetable\ continued from previous page}} 
\vspace{0.1cm}
\\
\hline
 & \multicolumn{3}{c}{ASFM} & \multicolumn{3}{c}{CRN} & \multicolumn{3}{c}{ECG} & \multicolumn{3}{c|}{FoldingNet} & \multicolumn{3}{c}{PPCL} \\ \cline{2-16} 
\multirow{-2}{*}{\begin{tabular}[c]{@{}c@{}}F-Score/CD-$\ell1$($\times$\\$10^{3}$)/CD-$\ell2$($\times10^{3}$)\end{tabular}} & Simple & Moderate & Hard & Simple & Moderate & Hard & Simple & Moderate & Hard & Simple & Moderate & Hard & Simple & Moderate & Hard \\ \hline
\endhead
\hline
\endfoot
\endlastfoot
\rowcolor[HTML]{EFEFEF} 
\cellcolor[HTML]{EFEFEF} & 0.184 & 0.183 & 0.167 & 0.137 & 0.139 & 0.139 & 0.500 & 0.489 & 0.391 & 0.065 & 0.065 & 0.053 & 0.408 & 0.387 & 0.306 \\
\rowcolor[HTML]{EFEFEF} 
\cellcolor[HTML]{EFEFEF} & 21.686 & 22.636 & 26.810 & 23.643 & 24.563 & 28.981 & 14.139 & 15.813 & 20.925 & 28.611 & 29.084 & 33.975 & 10.242 & 12.212 & 17.783 \\
\rowcolor[HTML]{EFEFEF} 
\multirow{-3}{*}{\cellcolor[HTML]{EFEFEF}Pillow} & 1.556 & 1.727 & 2.579 & 1.689 & 1.894 & 2.985 & 0.797 & 1.037 & 1.916 & 2.324 & 2.463 & 3.648 & 0.537 & 0.826 & 1.943 \\
 & 0.148 & 0.150 & 0.140 & 0.102 & 0.105 & 0.108 & 0.417 & 0.424 & 0.365 & 0.054 & 0.051 & 0.041 & 0.358 & 0.353 & 0.303 \\
 & 24.617 & 26.000 & 30.163 & 28.580 & 30.435 & 35.035 & 17.179 & 19.781 & 25.216 & 32.414 & 34.084 & 39.060 & 11.585 & 13.774 & 18.858 \\
\multirow{-3}{*}{Printer} & 2.158 & 2.578 & 3.541 & 2.621 & 3.321 & 4.689 & 1.286 & 1.942 & 3.060 & 3.303 & 3.908 & 5.140 & 0.685 & 1.223 & 2.364 \\
\rowcolor[HTML]{EFEFEF} 
\cellcolor[HTML]{EFEFEF} & 0.085 & 0.091 & 0.094 & 0.068 & 0.073 & 0.080 & 0.382 & 0.396 & 0.348 & 0.032 & 0.030 & 0.024 & 0.279 & 0.259 & 0.214 \\
\rowcolor[HTML]{EFEFEF} 
\cellcolor[HTML]{EFEFEF} & 32.204 & 33.850 & 39.939 & 36.891 & 40.285 & 49.334 & 20.047 & 23.604 & 31.438 & 39.057 & 41.653 & 49.682 & 14.067 & 18.102 & 26.183 \\
\hline
\end{longtable}}

\newpage
\setlength\tabcolsep{2pt}{
\tiny
\begin{longtable}[c]{cccccccccccccccc|ccc}
\caption{Quantitative completion results of GRNet, PCN, TopNet, PoinTr, Snowflake, and MMPT on ShapeNetUnseen21.}
\label{21}
\\
\hline
 & \multicolumn{3}{c}{GRNet~\citep{xie2020grnet}} & \multicolumn{3}{c}{PCN~\citep{yuan2018pcn}} & \multicolumn{3}{c}{TopNet~\citep{tchapmi2019topnet}} & \multicolumn{3}{c}{PoinTr~\citep{yu2021pointr}} & \multicolumn{3}{c|}{SnowflakeNet~\citep{xiang2021snowflakenet}} & \multicolumn{3}{c}{MMPT} \\ \cline{2-19} 
\multirow{-2}{*}{\begin{tabular}[c]{@{}c@{}}F-Score/CD-$\ell1$($\times$\\$10^{3}$)/CD-$\ell2$($\times10^{3}$)\end{tabular}} & Simple & Moderate & Hard & Simple & Moderate & Hard & Simple & Moderate & Hard & Simple & Moderate & Hard & Simple & Moderate & Hard & Simple & Moderate & Hard \\ \hline
\endfirsthead
\multicolumn{19}{c}%
{{\bfseries Table \thetable\ continued from previous page}} 
\vspace{0.1cm}
\\
\hline
 & \multicolumn{3}{c}{GRNet~\citep{xie2020grnet}} & \multicolumn{3}{c}{PCN~\citep{yuan2018pcn}} & \multicolumn{3}{c}{TopNet~\citep{tchapmi2019topnet}} & \multicolumn{3}{c}{PoinTr~\citep{yu2021pointr}} & \multicolumn{3}{c|}{SnowflakeNet~\citep{xiang2021snowflakenet}} & \multicolumn{3}{c}{MMPT} \\ \cline{2-19} 
\multirow{-2}{*}{\begin{tabular}[c]{@{}c@{}}F-Score/CD-$\ell1$($\times$\\$10^{3}$)/CD-$\ell2$($\times10^{3}$)\end{tabular}} & Simple & Moderate & Hard & Simple & Moderate & Hard & Simple & Moderate & Hard & Simple & Moderate & Hard & Simple & Moderate & Hard & Simple & Moderate & Hard \\ \hline
\endhead
\hline
\endfoot
\endlastfoot
\rowcolor[HTML]{EFEFEF} 
\cellcolor[HTML]{EFEFEF} & 0.169 & 0.165 & 0.163 & 0.097 & 0.095 & 0.081 & 0.077 & 0.071 & 0.058 & 0.407 & 0.396 & 0.338 & 0.358 & 0.316 & 0.246 & 0.408 & 0.387 & 0.306 \\
\rowcolor[HTML]{EFEFEF} 
\cellcolor[HTML]{EFEFEF} & 21.467 & 23.303 & 27.951 & 25.524 & 26.248 & 30.591 & 26.656 & 27.454 & 32.671 & 13.299 & 14.993 & 21.446 & 14.297 & 16.828 & 23.184 & 10.242 & 12.212 & 17.783 \\
\rowcolor[HTML]{EFEFEF} 
\multirow{-3}{*}{\cellcolor[HTML]{EFEFEF}Pillow} & 1.380 & 1.725 & 2.854 & 1.920 & 2.084 & 3.084 & 2.051 & 2.218 & 3.429 & 0.711 & 0.986 & 2.193 & 0.722 & 1.051 & 2.126 & 0.537 & 0.826 & 1.943 \\
 & 0.137 & 0.136 & 0.132 & 0.068 & 0.066 & 0.057 & 0.059 & 0.054 & 0.045 & 0.377 & 0.374 & 0.337 & 0.297 & 0.273 & 0.229 & 0.358 & 0.353 & 0.303 \\
 & 23.532 & 26.118 & 31.604 & 31.257 & 32.947 & 38.031 & 30.169 & 31.720 & 35.986 & 14.901 & 17.194 & 22.927 & 15.948 & 18.727 & 24.674 & 11.585 & 13.774 & 18.858 \\
\multirow{-3}{*}{Printer} & 1.643 & 2.393 & 3.839 & 3.123 & 3.755 & 5.214 & 2.787 & 3.280 & 4.413 & 0.960 & 1.621 & 2.856 & 0.910 & 1.533 & 2.718 & 0.685 & 1.223 & 2.364 \\
\rowcolor[HTML]{EFEFEF} 
\cellcolor[HTML]{EFEFEF} & 0.104 & 0.103 & 0.104 & 0.044 & 0.043 & 0.040 & 0.031 & 0.027 & 0.024 & 0.341 & 0.332 & 0.298 & 0.220 & 0.200 & 0.166 & 0.279 & 0.259 & 0.214 \\
\rowcolor[HTML]{EFEFEF} 
\cellcolor[HTML]{EFEFEF} & 27.482 & 32.697 & 42.477 & 41.716 & 44.216 & 52.634 & 37.677 & 39.941 & 46.729 & 18.083 & 22.478 & 33.008 & 20.919 & 25.513 & 35.215 & 14.067 & 18.102 & 26.183 \\
\hline
\end{longtable}}

\end{document}